\documentclass{article}

\usepackage{arxiv}

\usepackage[utf8]{inputenc} % allow utf-8 input
\usepackage[T1]{fontenc}    % use 8-bit T1 fonts
\usepackage{hyperref}       % hyperlinks
\usepackage{url}            % simple URL typesetting
\usepackage{booktabs}       % professional-quality tables
\usepackage{amsfonts}       % blackboard math symbols
\usepackage{nicefrac}       % compact symbols for 1/2, etc.
\usepackage{microtype}      % microtypography
\usepackage{lipsum}
\usepackage{fancyhdr}       % header
\usepackage{graphicx}       % graphics
\graphicspath{{media/}}     % organize your images and other figures under media/ folder

\usepackage{algorithm}
\usepackage{algorithmic}
\usepackage{amsmath}
\usepackage{amssymb}
\usepackage{amsthm}
\usepackage{array}
\usepackage{enumitem}
\usepackage{mathtools}
\usepackage{multirow}

\DeclareMathOperator*{\argmin}{arg\,min}

\title{
    Curved Representation Space of Vision Transformers
}

\author{
    Juyeop Kim,
    Junha Park,
    Songkuk Kim,
    Jong-Seok Lee\\
    Yonsei University, Korea\\
    \texttt{
    \{juyeopkim, junha.park, songkuk, jong-seok.lee\}@yonsei.ac.kr
    }\\
}

\begin{document}
\maketitle

\begin{abstract}
Neural networks with self-attention (a.k.a. Transformers) like ViT and Swin have emerged as a better alternative to traditional convolutional neural networks (CNNs).
However, our understanding of how the new architecture works is still limited.
In this paper, we focus on the phenomenon that Transformers show higher robustness against corruptions than CNNs, while not being overconfident.
This is contrary to the intuition that robustness increases with confidence.
We resolve this contradiction by empirically investigating how the output of the penultimate layer moves in the representation space as the input data moves linearly within a small area.
In particular, we show the following.
(1) While CNNs exhibit fairly linear relationship between the input and output movements, Transformers show nonlinear relationship for some data.
For those data, the output of Transformers moves in a curved trajectory as the input moves linearly.
(2) When a data is located in a curved region, it is hard to move it out of the decision region since the output moves along a curved trajectory instead of a straight line to the decision boundary, resulting in high robustness of Transformers.
(3) If a data is slightly modified to jump out of the curved region, the movements afterwards become linear and the output goes to the decision boundary directly.
In other words, there does exist a decision boundary near the data, which is hard to find only because of the curved representation space.
This explains the underconfident prediction of Transformers.
Also, we examine mathematical properties of the attention operation that induce nonlinear response to linear perturbation.
Finally, we share our additional findings, regarding  what contributes to the curved representation space of Transformers, and how the curvedness evolves during training.
\end{abstract}

\section{Introduction}

Self-attention-based neural network architectures, including Vision Transformers~\cite{dosovitskiy2021vit}, Swin Transformers~\cite{liu2021swin}, etc. (hereinafter referred to as Transformers), have shown to outperform traditional convolutional neural networks (CNNs) in various computer vision tasks.
The success of the new architecture has prompted a question, how Transformers work, especially compared to CNNs, which would also shed light on deeper understanding of CNNs and eventually neural networks.

In addition to the improved task performance (e.g., classification accuracy) compared to CNNs, Transformers also show desirable characteristics in other aspects.
It has been shown that Transformers are more robust to adversarial perturbations than CNNs~\cite{bai2021robust1, naseer2021robust2, paul2022robust3}.
Moreover, Transformers are reported not overconfident in their predictions unlike CNNs~\cite{minderer2021revisiting} (and we show that Transformers are actually underconfident in this paper).

The high robustness, however, does not comport with underconfidence.
Intuitively, a data that is correctly classified by a model with lower confidence is likely to be located closer to the decision boundary (see Appendix for detailed discussion).
Then, a smaller amount of perturbation would move the data out of the decision region, which translates into lower robustness of the model.
However, the previous results claim the opposite.

To mitigate the contradiction of robustness and underconfidence, this paper presents our empirical study to explore the representation space of Transformers and CNNs.
More specifically, we focus on the \textit{linearity of the models}, i.e., the change of the output feature (which is simply referred to as output in this paper) with respect to the linear change of the input data.
It is known that adversarial examples are a result of models being too linear, based on which the fast gradient sign method (FGSM) was introduced to show that deep neural networks can be easily fooled~\cite{goodfellow2015fgsm}.
Motivated by this, we examine the input-output relationship of Transformers through the course that the input is gradually perturbed along the direction determined by FGSM.

% fig:visualize
\begin{figure}[!t]
    \centering
    \small

    \begin{tabular}{ccc}
        \includegraphics[width=.30\linewidth]{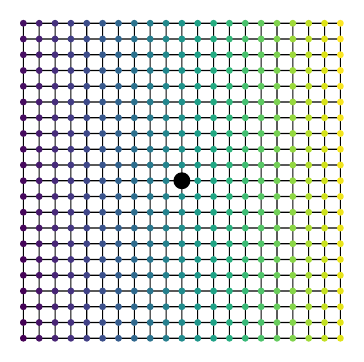} &
        \includegraphics[width=.30\linewidth]{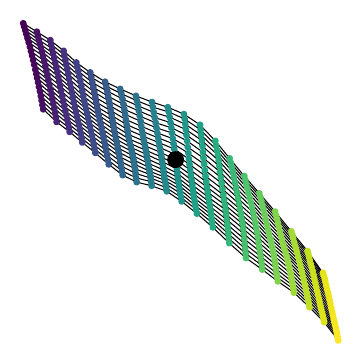} &
        \includegraphics[width=.30\linewidth]{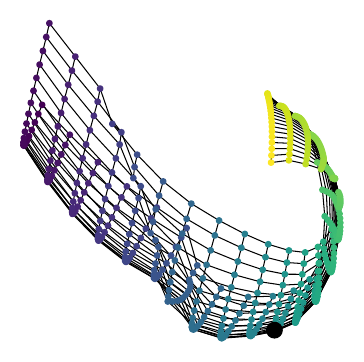} \\
        (a) Input space &
        (b) Representation space &
        (c) Representation space \\
        &
        (ResNet50) &
        (Swin-T)
    \end{tabular}

    \caption{
        2D projected movements of (a) the data (black dot) in the input space and corresponding output features in the representation space for (b) ResNet50 and (c) Swin-T.
    }\label{fig:visualize}
\end{figure}

Fig.~\ref{fig:visualize} visualizes the representation spaces of CNNs and Transformers comparatively (see Appendix for implementation details).
An image data from ImageNet~\cite{russakovsky15imagenet}, marked with the black dot in Fig.~\ref{fig:visualize}a, is gradually modified by a fixed amount along two mutually orthogonal directions.
The corresponding outputs of ResNet50~\cite{he2016resnet} and Swin-T~\cite{liu2021swin} are obtained, which are shown after two-dimensional projection in Figs.~\ref{fig:visualize}b and \ref{fig:visualize}c, respectively.
While the gradual changes of the input produce almost linear changes in the output of ResNet50, the output trajectory of Swin-T is nonlinear around the original output (and then becomes linear when the change of the output is large), i.e., the representation space is locally \textit{curved}.
We empirically show that this curved representation space results in the aforementioned contradiction.

Our main research questions and findings can be summarized as follows.

1. \textit{How does the representation space of Transformers look like?}
To answer this, we analyze the movement of the penultimate layer's output with respect to the linear movement of the input.
We use the adversarial gradient produced by FGSM~\cite{goodfellow2015fgsm} as the direction of movement in the input space, to investigate the linearity of the feature space of the models.
We find that the directions of successive movements of the output significantly change in the case of Transformers unlike CNNs, indicating that \textbf{the representation space of Transformers is locally curved}.

2. \textit{What makes Transformers robust to input perturbation?}
\textbf{We find that the curved regions in the representation space account for the robustness of Transformers}.
When a data is located in a curved region, a series of linear perturbations to the input move the output point along a curved trajectory.
This makes it hard to move the data out of its decision region along a short and straight line, which explains high robustness of Transformers for the data.

3. \textit{Then, why is the prediction of Transformers underconfident?}
Although it takes many steps to escape from a curved decision region and reach a decision boundary, we find that a decision boundary is actually located closely to the original output.
We demonstrate a simple trick to reach the decision boundary quickly.
I.e., when a small amount of random noise is added to the input data, its output can jump out of the locally curved region and arrive at a linear region, from which a closely located decision boundary can be reached by adding only a small amount of perturbation. This reveals that \textbf{the decision boundary exists near the original data in the representation space, which explains the underconfident predictions of Transformers}.

We also present additional observations examining what contributes to the curved representation space of Transformers and when the curvedness is formed during training.

\section{Related work}

Since the first application of the self-attention mechanism to vision tasks \cite{dosovitskiy2021vit}, a number of studies have shown that the models built with traditional convolutional layers are outperformed by Transformers utilizing self-attention layers in terms of task performance \cite{liu2021swin,chu2021twins,huang2021shuffle,li2021localvit,touvron2021deit,wang2021pyramid,xiao2021early,yang2021focal,yuan2021tokens,liu2022swin}.
There have been efforts to compare CNNs and Transformers in various aspects.
Empirical studies show that Transformers have higher adversarial robustness than CNNs \cite{paul2022robust3,naseer2021robust2,aldahdooh2021reveal,bhojanapalli2021understanding}, which seems to be due to the reliance of Transformers on lower frequency information than CNNs \cite{park2021how,benz2021adversarial}.
Other studies conclude that Transformers are calibrated better than CNNs yielding overconfident predictions \cite{guo2017calibration,thulasidasan2019mixup,wen2021combining,minderer2021revisiting}.
However, there has been no clear explanation encompassing both higher robustness and lower confidence of Transformers.

Understanding how neural networks work has been an important research topic.
A useful way for this is to investigate the input-output mapping formed by a model.
Since models with piecewise linear activation functions (e.g., ReLU) implement piecewise linear mappings, several studies investigate the characteristics of linear regions, e.g., counting the number of linear regions as a measure of model expressivity (or complexity) \cite{montufar2014number,hanin2019complexity,hanin2019deep,telgarsky2015representation,serra2018bounding,raghu2017expressive} and examining local properties of linear regions \cite{zhang20empirical}.
Some studies examine the length of the output curve for a given unit-length input \cite{raghu2017expressive,price2019trajectory,hanin2022deep}.
There also exist some works that relate the norm of the input-output Jacobian matrix to generalization performance \cite{sokolic2017robust,novak2018sensitivity}.
However, the input-output relationship of Transformers has not been explored previously, which is focused in this paper.

\section{On the ostensible contradiction of high robustness and underconfidence}

\subsection{Model calibration}

It is desirable that a trained classifier is well-calibrated by making prediction with reasonable certainty, e.g., for data that a classifier predicts with confidence (i.e., probability of the predicted class) of 80\%, its accuracy should also be 80\% in average.
A common measure to evaluate model calibration is the expected calibration error (ECE) defined as \cite{naeini2015obtaining}
\begin{equation}
    \mathrm{ECE} = \sum_{i=1}^{K}{P(i)\cdot|o_{i}-e_{i}|},
    \label{eq:ece}
\end{equation}
where $K$ is the number of bins of confidence, $P(i)$ is the fraction of data falling into bin $i$, $o_i$ is the accuracy of the data in bin $i$, and $e_i$ is the average confidence of the data in bin $i$.
One limitation of ECE is that it does not distinguish between overconfidence and underconfidence because the sign of the difference between the accuracy and the confidence is ignored.
Therefore, we define signed ECE (sECE) to augment ECE, as follows.
\begin{equation}
    \mathrm{sECE} = \sum_{i=1}^{K}{P(i)\cdot(o_{i}-e_{i})}.
    \label{eq:sece}
\end{equation}
An overconfident model will have higher confidence than accuracy, resulting in a negative sECE value.
An underconfident model, in contrast, will show a positive value of sECE.

We compare the calibration of CNNs, including ResNet50 \cite{he2016resnet} and MobileNetV2 \cite{sandler2018mobilenetv2, howard2019mobilenetv3}, and Transformers, including ViT-B/16 \cite{dosovitskiy2021vit} and Swin-T \cite{liu2021swin}, on the ImageNet validation set using ECE and sECE in Fig.~\ref{fig:calibration} (see Fig.~\ref{fig:calibration_2} in Appendix for the results of other models).
CNNs show negative ECE values and bar plots below the 45$^\circ$ line, indicating overconfidence in prediction, which is consistent with the previous studies \cite{guo2017calibration}.
On the other hand, Transformers are underconfident, showing positive sECE and bar plots over the 45$^\circ$ line.
This comparison result is interesting: Transformers reportedly show higher classification accuracy than CNNs, but in fact with lower confidence.

% fig:calibration
\begin{figure}[!t]
    \centering
    \small

    \begin{tabular}{cccc}
        \includegraphics[width=.20\linewidth]{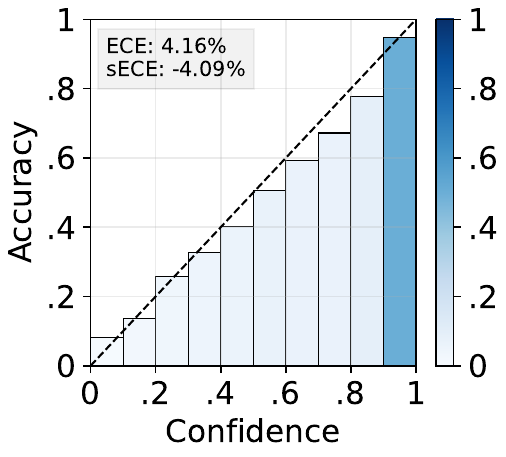} &
        \includegraphics[width=.20\linewidth]{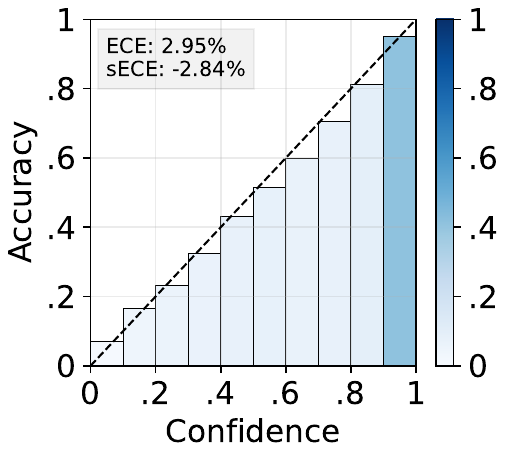} &
        \includegraphics[width=.20\linewidth]{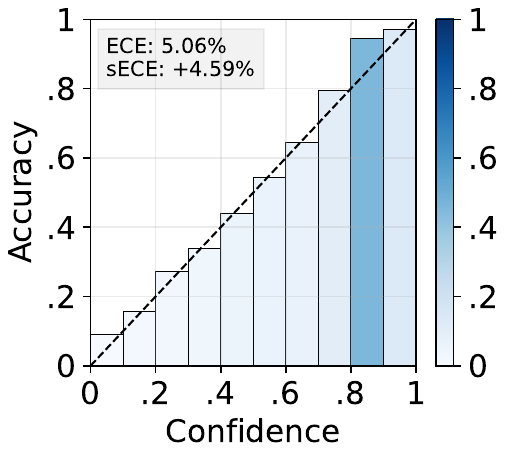} &
        \includegraphics[width=.20\linewidth]{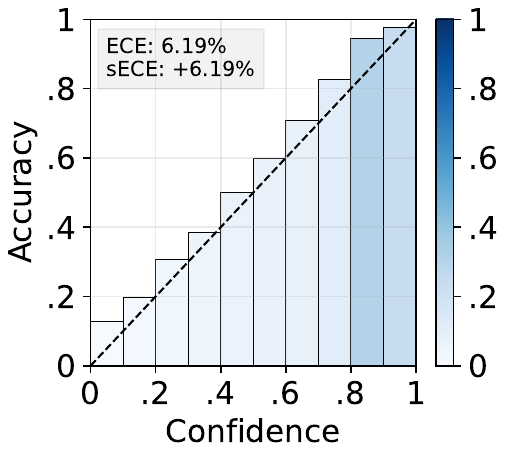} \\
        (a) ResNet50 &
        (b) MobileNetV2 &
        (c) ViT-B/16 &
        (d) Swin-T
    \end{tabular}

    \caption{
        Reliability diagrams of CNNs and Transformers.
        Transparency of bars represent the ratio of the number of data in each confidence bin.
        ECE and sECE values are also shown in each case.
    }\label{fig:calibration}
\end{figure}

% fig:fgsm-travel
\begin{figure}[!t]
    \centering
    \small

    \begin{tabular}{cccc}
        \includegraphics[width=.20\linewidth]{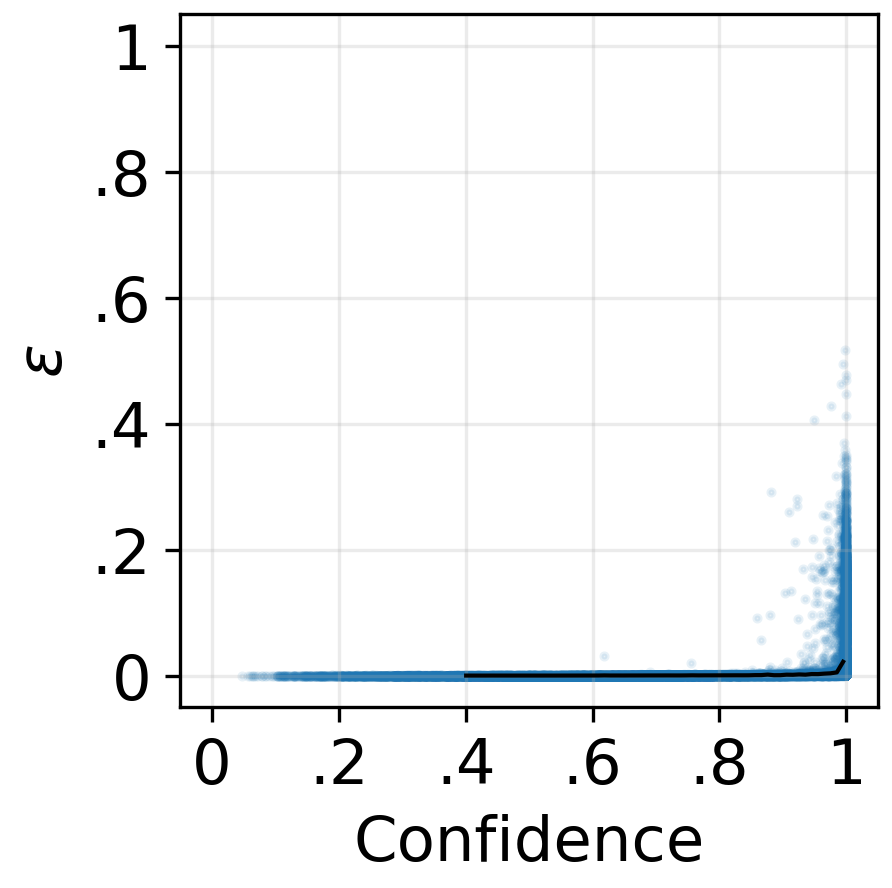} &
        \includegraphics[width=.20\linewidth]{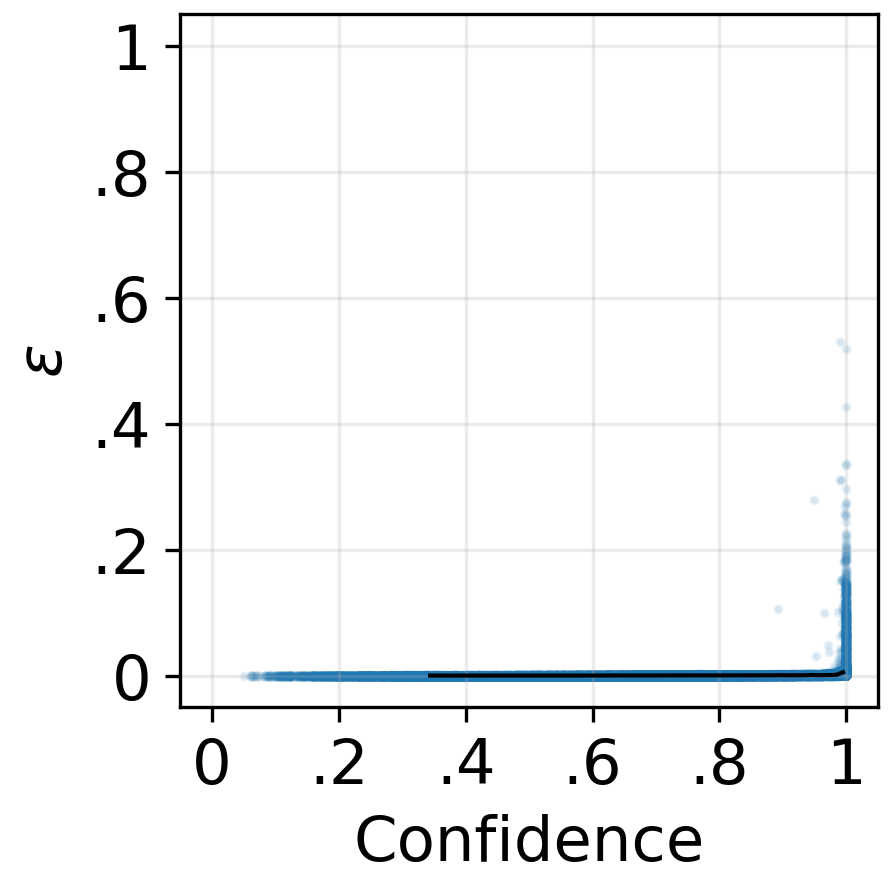} &
        \includegraphics[width=.20\linewidth]{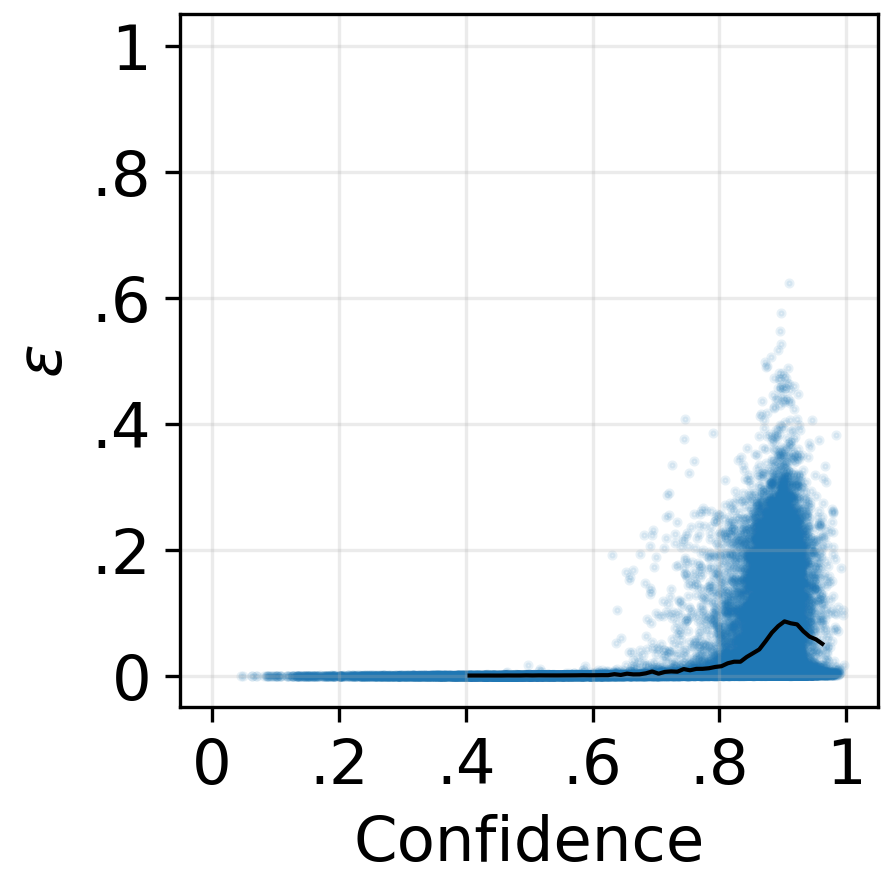} &
        \includegraphics[width=.20\linewidth]{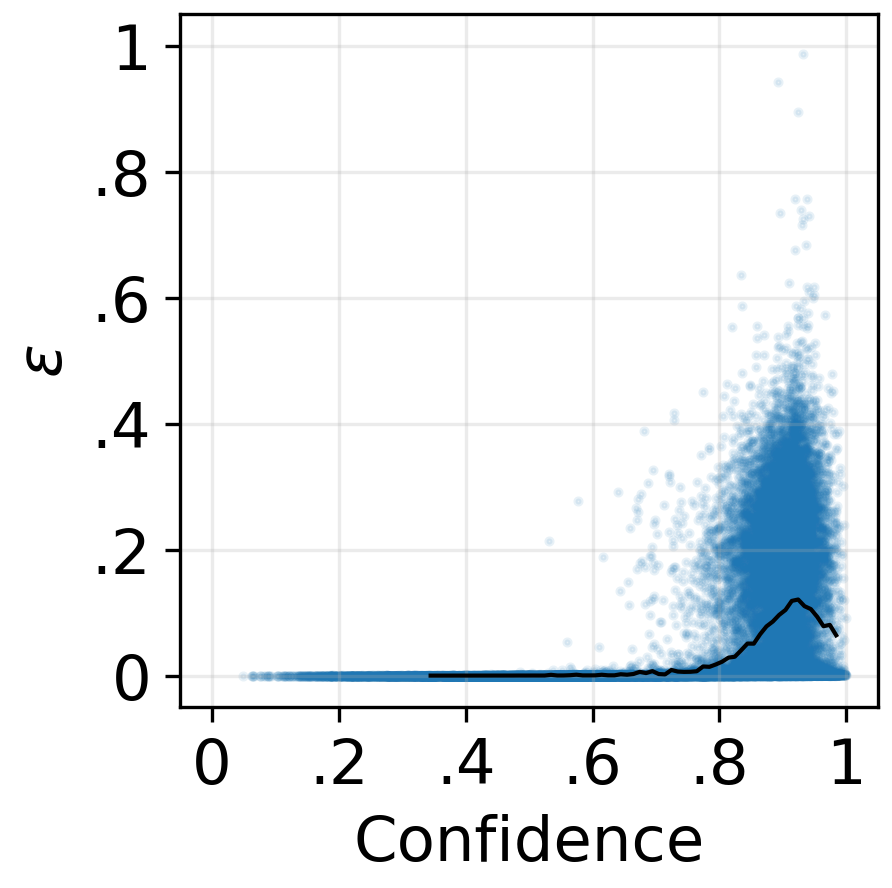} \\
        \hspace{5.0mm} (a) ResNet50 &
        \hspace{5.0mm} (b) MobileNetV2 &
        \hspace{5.0mm} (c) ViT-B/16 &
        \hspace{5.0mm} (d) Swin-T
    \end{tabular}

    \caption{
        Lengths ($\epsilon$) of the travel to decision boundaries with respect to the confidence for the ImageNet validation data.
        Black lines represent average values.
    }\label{fig:fgsm-travel}
\end{figure}

\subsection{Passage to decision boundary}

It is a common intuition that if a model classifies a data with low confidence, the data is likely to be located near a decision boundary (see Appendix for detailed discussion).
Based on the above results, therefore, the decision boundaries of Transformers are assumed to be formed near the data compared to CNNs.
To validate this, we formulate a procedure to examine the distance to a decision boundary from a data through a linear travel.
Concretely, we aim to solve the following optimization problem:
\begin{equation}
    \argmin_{\epsilon} ~ \mathcal{C}(\mathrm{\mathbf{x}}^{\prime}) \neq y, ~ ~ ~ ~ ~ ~ ~ ~ ~ ~
    \mathrm{\mathbf{x}}^{\prime} = \mathrm{\mathbf{x}} + \epsilon \cdot \mathrm{\mathbf{d}},
    \label{eq:travel}
\end{equation}
where $\mathrm{\mathbf{x}}$ is the input data, $y$ is the true class label of $\mathrm{\mathbf{x}}$, $\mathcal{C}$ is the classifier, $\mathrm{\mathbf{d}}$ is the travel direction, $\epsilon$ is a positive real number indicating the travel length, and $\mathrm{\mathbf{x}}^{\prime}$ is the traveled result of $\mathrm{\mathbf{x}}$.
We set the travel direction $\mathrm{\mathbf{d}}$ as the adversarial gradient produced by FGSM, i.e.,
\begin{equation}
    \mathrm{\mathbf{d}} = \mathrm{sign}(\nabla_{\mathrm{\mathbf{x}}} J(\mathcal{C}(\mathrm{\mathbf{x}}), y)),
    \label{eq:fgsm}
\end{equation}
where $J$ is the classification loss function (i.e., cross-entropy).
Note that $\|\mathbf{d}\|_{2}=\sqrt{D}$, where $D$ is the dimension of $\mathrm{\mathbf{x}}$.
Refer to Algorithm~\ref{alg:travel} in Appendix for the detailed procedure to solve the optimization problem in Eq.~\ref{eq:travel}.

Fig.~\ref{fig:fgsm-travel} shows the obtained values of $\epsilon$ with respect to the confidence values for the ImageNet validation data (see Fig.~\ref{fig:fgsm-travel_2} in Appendix for the results of other models).
On the contrary to our expectation, decision boundaries seem to be located farther from the data in the input space for Transformers than CNNs.
This contradiction is resolved in the following section.

\section{Resolving the contradiction}

\subsection{Shape of representation space}

As mentioned in the \textbf{Introduction}, the FGSM attack was first introduced to show that the linearity of a model causes its vulnerability to adversarial perturbations~\cite{goodfellow2015fgsm}.
To resolve the contradiction between high robustness (a large distance to the decision boundary) and underconfidence (a small distance to the decision boundary) of Transformers in the previous section, therefore, we examine the degree of linearity of the input-output relationship, i.e., how linear movements in the input space appear in the representation space of Transformers.

We divide the travel into $N$ steps as
\begin{equation}
    \mathrm{\mathbf{x}}^{(n)} = \mathrm{\mathbf{x}}^{(0)} + n \cdot \frac{\epsilon}{N}\mathrm{\mathbf{d}}, ~ ~ ~ ~ ~
    (n=0, 1, \cdots, N)
\end{equation}
where $\mathrm{\mathbf{x}}^{(0)} = \mathrm{\mathbf{x}}$ and $\mathrm{\mathbf{x}}^{(N)}$
are the initial and final data points, respectively.
For each $\mathrm{\mathbf{x}}^{(n)}$, we obtain its output feature at the penultimate layer, which is denoted as $\mathrm{\mathbf{z}}^{(n)}$.
Unlike the travel in the input space, the magnitude and direction of the travel appearing in the representation space may change at each step.
Thus, the movement at step $n$ is defined as
\begin{equation}
    \mathrm{\mathbf{d}}^{(n)}_{\mathrm{\mathbf{z}}} = \mathrm{\mathbf{z}}^{(n)} - \mathrm{\mathbf{z}}^{(n-1)},
\end{equation}
from which the magnitude ($\omega^{(n)}$) and relative direction ($\theta^{(n)}$) are obtained as
\begin{equation}
    \omega^{(n)} =
        \|\mathrm{\mathbf{d}}^{(n)}_{\mathrm{\mathbf{z}}}\|, ~ ~ ~ ~ ~ ~ ~ ~ ~ ~
    \theta^{(n)} =
        \cos^{-1} \left(\frac{
            \mathrm{\mathbf{d}}^{(n)}_\mathrm{\mathbf{z}} \cdot \mathrm{\mathbf{d}}^{(n+1)}_\mathrm{\mathbf{z}}
            }{
            \|\mathrm{\mathbf{d}}^{(n)}_\mathrm{\mathbf{z}}\| \|\mathrm{\mathbf{d}}^{(n+1)}_\mathrm{\mathbf{z}}\|
            }\right).
    \label{eq:distance_and_direction_change}
\end{equation}

% fig:illust
\begin{figure}[!t]
    \centering
    \small

    \begin{tabular}{cc}
        \hspace{-2.0mm} \includegraphics[width=.46\linewidth]{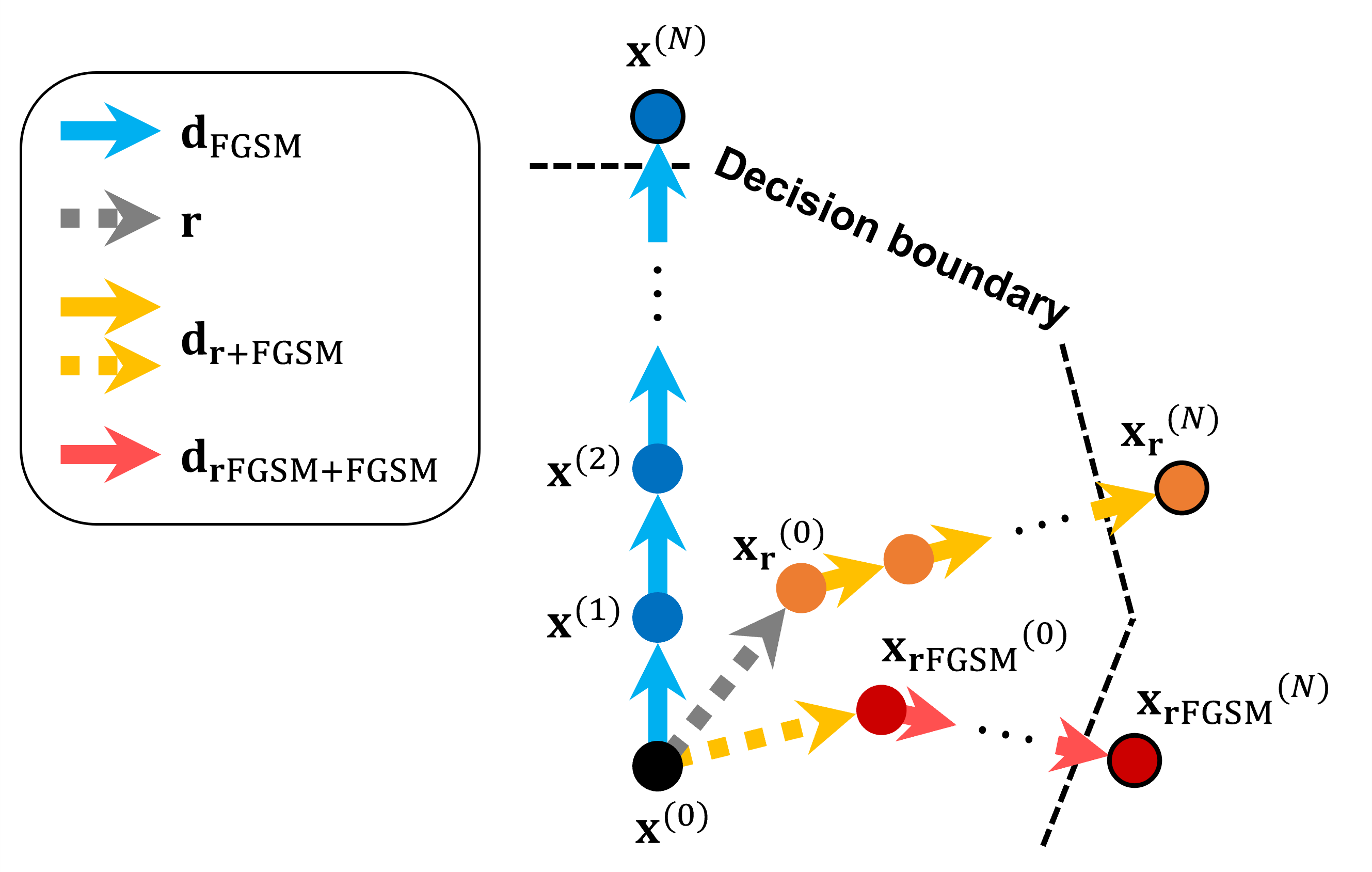} &
        \hspace{-5.0mm} \includegraphics[width=.45\linewidth]{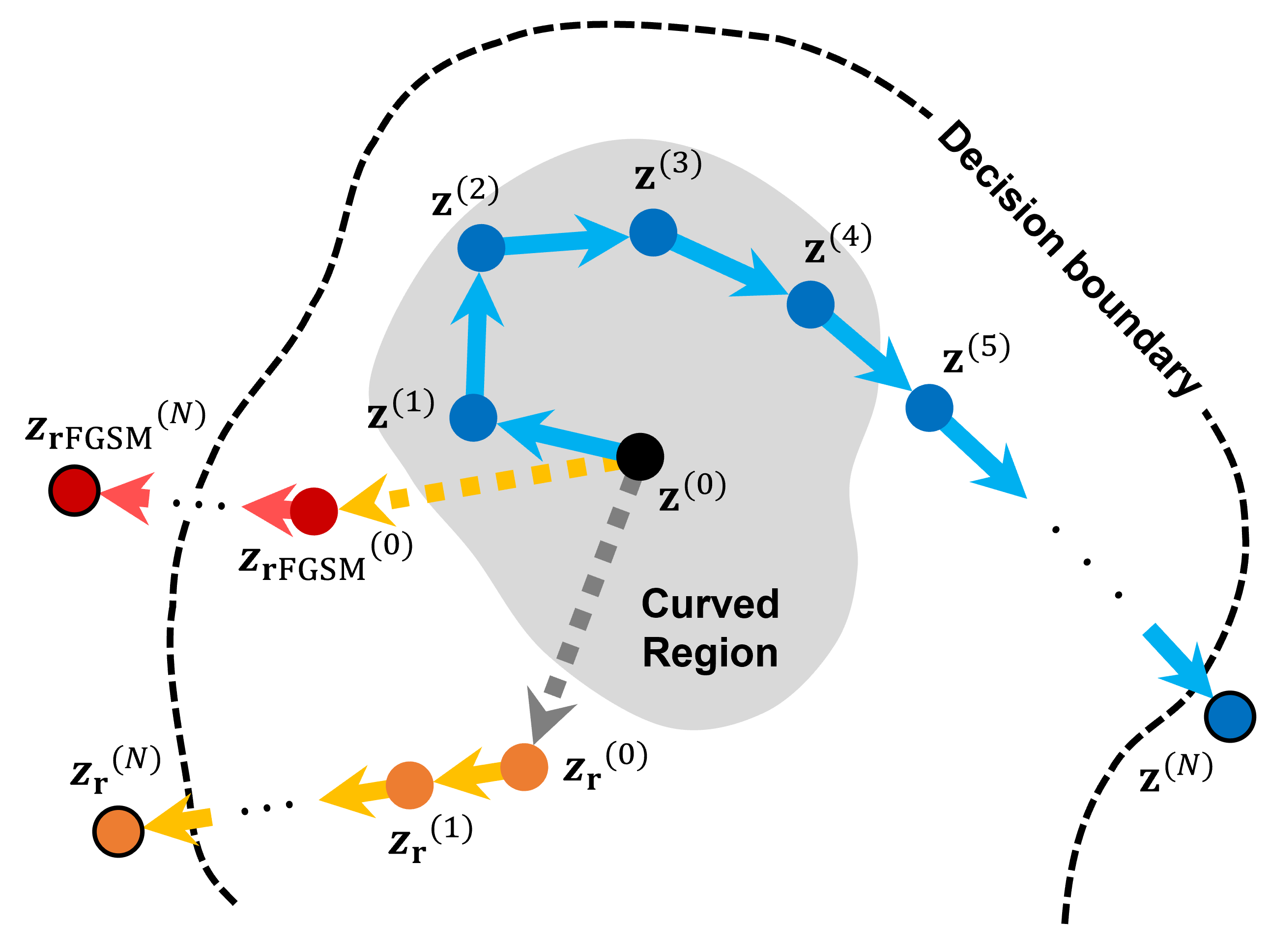} \\
        \hspace{-2.0mm} (a) Input space &
        \hspace{-5.0mm} (b) Representation space
    \end{tabular}

    \caption{
        Illustration of the input-output relationship of Transformers in terms of the trajectories in the input space and the representation space.
    }\label{fig:illust}
\end{figure}

We consider three different ways to determine $\mathrm{\mathbf{d}}$:
\begin{itemize}[leftmargin=*,align=left]
    \item $\mathrm{\mathbf{d}}_{\mathrm{FGSM}}$ (blue-colored trajectory in Fig.~\ref{fig:illust}):
    FGSM direction for $\mathrm{\mathbf{x}}^{(0)}$ (as in Eq.~\ref{eq:fgsm}).
    
    \item $\mathrm{\mathbf{d}}_{\mathrm{\mathbf{r}} + \mathrm{FGSM}}$ (yellow-colored trajectory in Fig.~\ref{fig:illust}): FGSM direction determined for the randomly perturbed data $\mathrm{\mathbf{x}}_{\mathrm{\mathbf{r}}}^{(0)} = \mathrm{\mathbf{x}}^{(0)} + \epsilon_{\mathrm{\mathbf{r}}} \cdot \mathrm{\mathbf{r}}$, where $\mathrm{\mathbf{r}}$ is a random vector ($\|\mathrm{\mathbf{r}}\|_{2}=\sqrt{D}$) and $\epsilon_{\mathrm{\mathbf{r}}}$ controls the amount of this ``random jump.''
    
    \item $\mathrm{\mathbf{d}}_{\mathrm{\mathbf{r}FGSM} + \mathrm{FGSM}}$ (red-colored trajectory in Fig.~\ref{fig:illust}): FGSM direction determined for the data perturbed in the direction of $\mathrm{\mathbf{d}}_{\mathrm{\mathbf{r}} + \mathrm{FGSM}}$, i.e., $\mathrm{\mathbf{x}}^{(0)}_{\mathrm{\mathbf{r}FGSM}} = \mathrm{\mathbf{x}}^{(0)} + \epsilon_{\mathrm{\mathbf{r}} + \mathrm{FGSM}} \cdot \mathrm{\mathbf{d}}_{\mathrm{\mathbf{r}} + \mathrm{FGSM}}$, where $\epsilon_{\mathrm{\mathbf{r}} + \mathrm{FGSM}}$ controls the amount of this jump.
\end{itemize}

% fig:footprint_csim
\begin{figure}[!t]
    \centering
    \small

    \begin{tabular}{ccccc}
        &
        \hspace{4.0mm} ResNet50 &
        \hspace{4.0mm} MobileNetV2 &
        \hspace{4.0mm} ViT-B/16 &
        \hspace{4.0mm} Swin-T \\

        \rotatebox[origin=l]{90}{\hspace{13.0mm}$\mathrm{\mathbf{d}_{FGSM}}$} &
        \includegraphics[width=.20\linewidth]{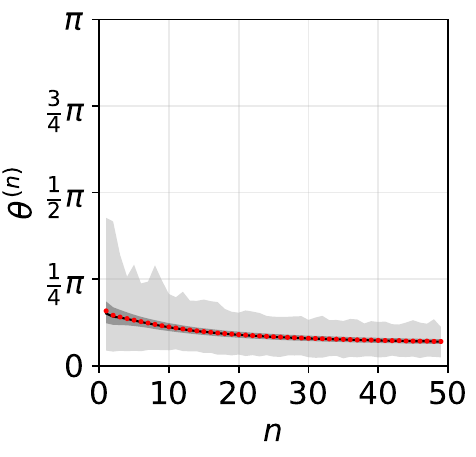} &
        \includegraphics[width=.20\linewidth]{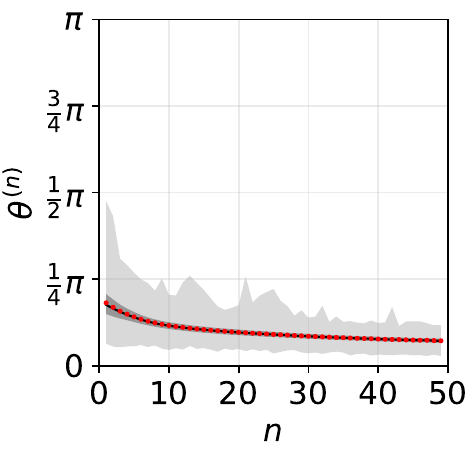} &
        \includegraphics[width=.20\linewidth]{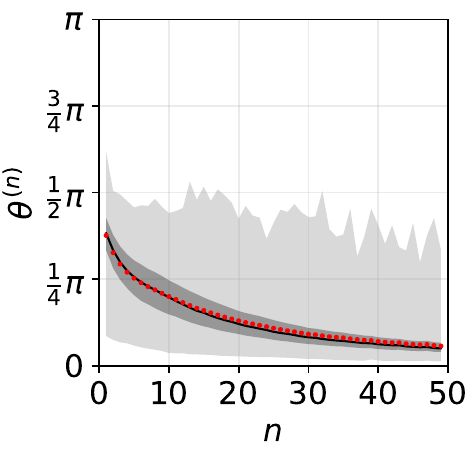} &
        \includegraphics[width=.20\linewidth]{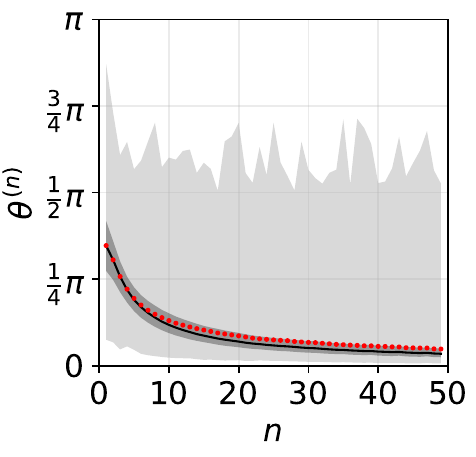} \\
        &
        \hspace{4.0mm} (a) &
        \hspace{4.0mm} (b) &
        \hspace{4.0mm} (c) &
        \hspace{4.0mm} (d) \\

        \rotatebox[origin=l]{90}{\hspace{11.0mm}$\mathrm{\mathbf{d}_{\mathbf{r}+FGSM}}$} &
        \includegraphics[width=.20\linewidth]{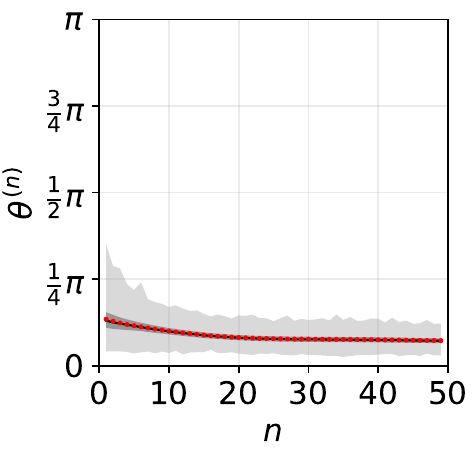} &
        \includegraphics[width=.20\linewidth]{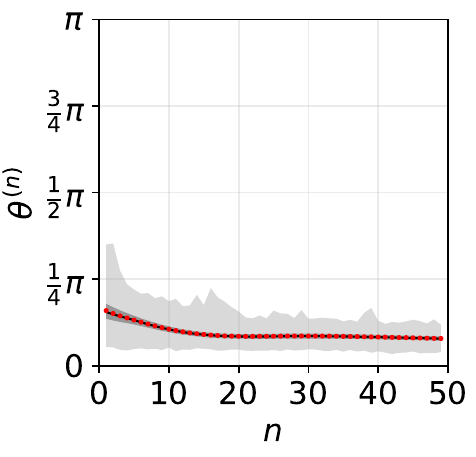} &
        \includegraphics[width=.20\linewidth]{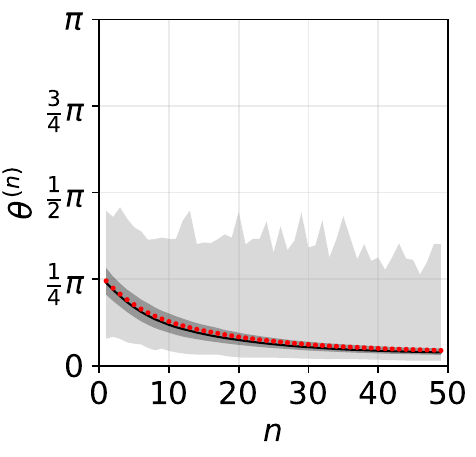} &
        \includegraphics[width=.20\linewidth]{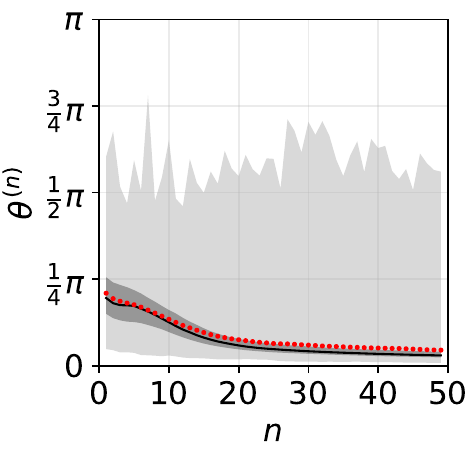} \\
        &
        \hspace{4.0mm} (e) &
        \hspace{4.0mm} (f) &
        \hspace{4.0mm} (g) &
        \hspace{4.0mm} (h)
    \end{tabular}

    \caption{
        Direction changes of output features with respect to the travel step ($n$).
        \textbf{Light-gray regions}: Range between the minimum and maximum values.
        \textbf{Dark-gray regions}: Range between the first quartile (Q1) and the third quartile (Q3).
        \textbf{Black lines}: Medians (Q2).
        \textbf{Red dots}: Mean values.
    }\label{fig:footprint_csim}
\end{figure}

Figs.~\ref{fig:footprint_csim}a-\ref{fig:footprint_csim}d show the direction changes in travel for ResNet50, MobileNetV2, ViT-B/16 and Swin-T when $\mathrm{\mathbf{d}} = \mathrm{\mathbf{d}}_{\mathrm{FGSM}}$, $\epsilon = .05$, and $N = 50$.
See Fig.~\ref{fig:footprint_around_csim} in Appendix for the results of other travel directions, which shows a similar trend.
For ResNet50 and MobileNetV2, the direction does not change much (Figs.~\ref{fig:footprint_csim}a and \ref{fig:footprint_csim}b), which in fact holds regardless of the travel direction in the input space (see Figs.~\ref{fig:footprint_around_csim}a and \ref{fig:footprint_around_csim}b in Appendix).
This indicates that \textbf{the input-output relationship of CNNs is fairly linear around the data}.
In contrast, ViT-B/16 and Swin-T shows locally nonlinear input-output relationship;
$\theta^{(n)}$ is significantly large in early steps of travel (Figs.~\ref{fig:footprint_csim}c and \ref{fig:footprint_csim}d), even for other travel directions in the input space (see Figs.~\ref{fig:footprint_around_csim}c and \ref{fig:footprint_around_csim}d in Appendix).
I.e., \textbf{Transformers generate \textit{nonlinear} response to linear perturbation and the representation space of Transformers is \textit{curved} around the data}.

Figs.~\ref{fig:footprint_csim}e-\ref{fig:footprint_csim}h show the direction changes in travel when $\mathrm{\mathbf{d}} = \mathrm{\mathbf{d}}_{\mathrm{\mathbf{r}} + \mathrm{FGSM}}$, with $\epsilon_{\mathrm{\mathbf{r}}} = .05$ except for ViT-B/16 using $\epsilon_{\mathrm{\mathbf{r}}} = .20$ (see Appendix for discussion).
It can be observed that the direction does not change much after the random jump.
I.e., \textbf{the curvedness of the representation space is localized around the data}.
Therefore, by making $\mathrm{\mathbf{x}}^{(0)}$ jump a certain distance in a random direction $\mathrm{\mathbf{r}}$, $\mathrm{\mathbf{z}}^{(0)}$ can pass over the curved region without meandering in the early steps and make linear movements afterwards ($\mathbf{z_r}^{(n)}$ in Fig.~\ref{fig:illust}b).

% fig:dist_to_db_hist
\begin{figure}[!t]
    \centering
    \small

    \begin{tabular}{cccc}
        \includegraphics[width=.20\linewidth]{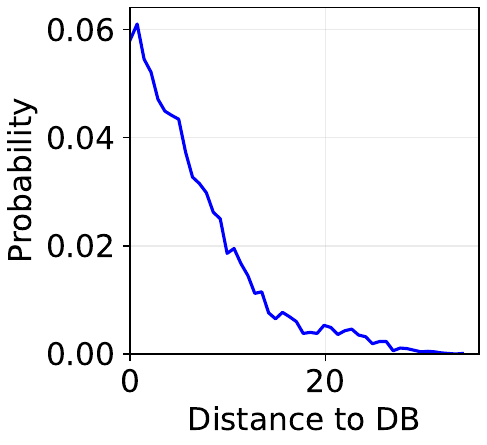} &
        \includegraphics[width=.20\linewidth]{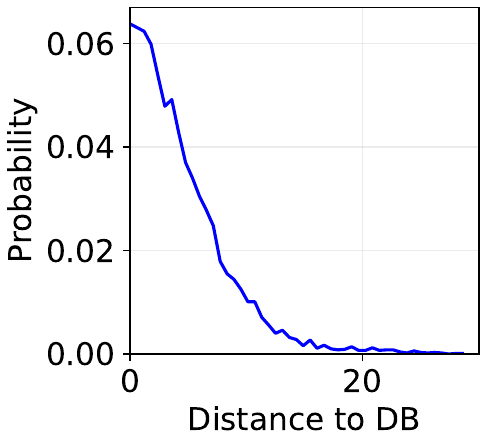} &
        \includegraphics[width=.20\linewidth]{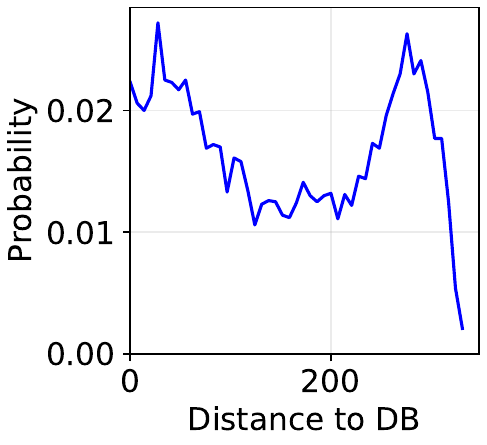} &
        \includegraphics[width=.20\linewidth]{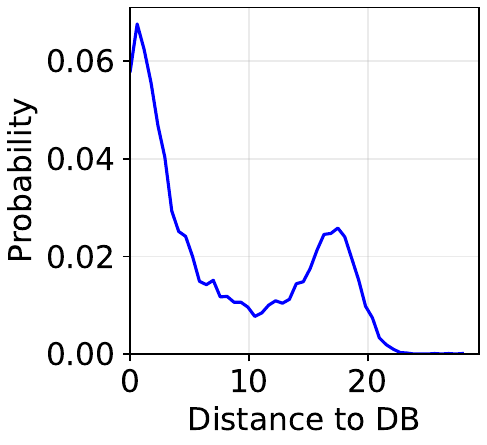} \\
        \hspace{7.0mm} (a) ResNet50 &
        \hspace{7.0mm} (b) MobileNetV2 &
        \hspace{7.0mm} (c) ViT-B/16 &
        \hspace{7.0mm} (d) Swin-T
    \end{tabular}

    \caption{
        Distribution of distance from the original output to the decision boundary (DB) in the representation space.
    }\label{fig:dist_to_db_hist}
\end{figure}

% fig:dist_to_db_scatter
\begin{figure}[!t]
    \centering
    \small

    \begin{tabular}{cc}
        \includegraphics[width=.229\linewidth]{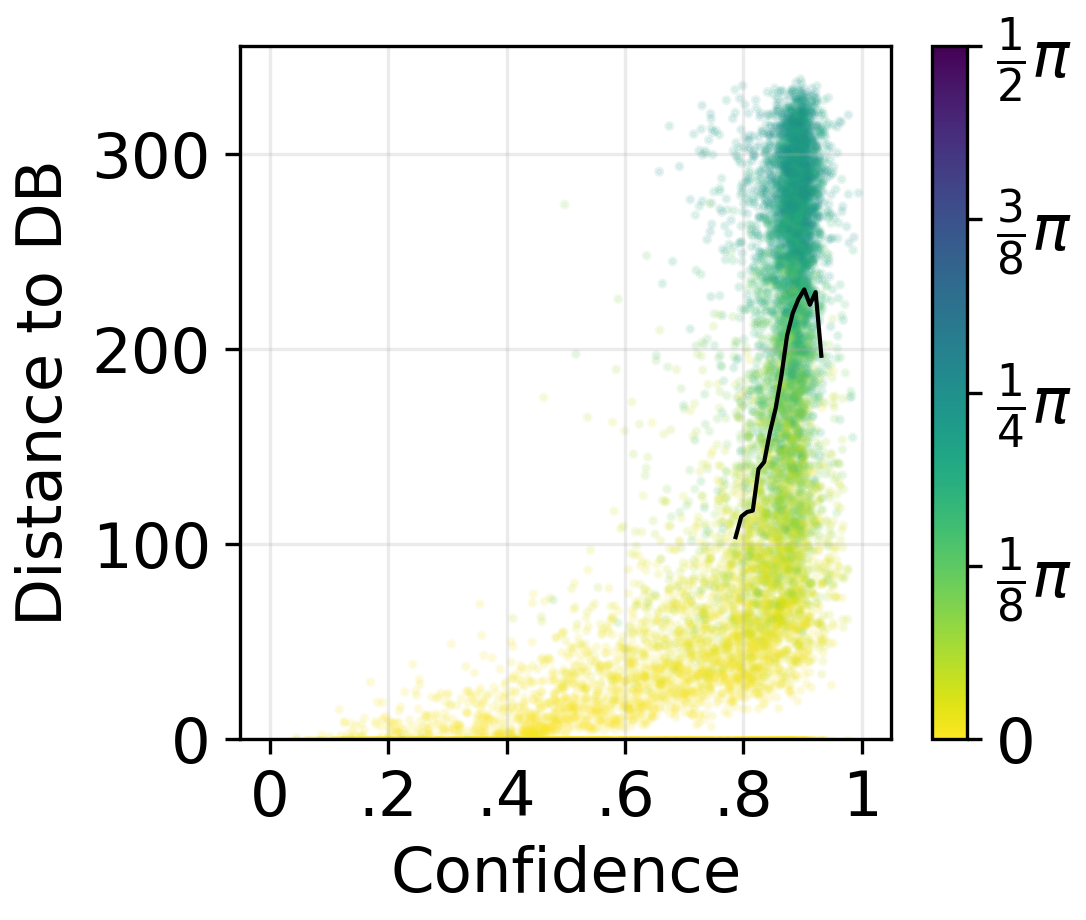} &
        \includegraphics[width=.22\linewidth]{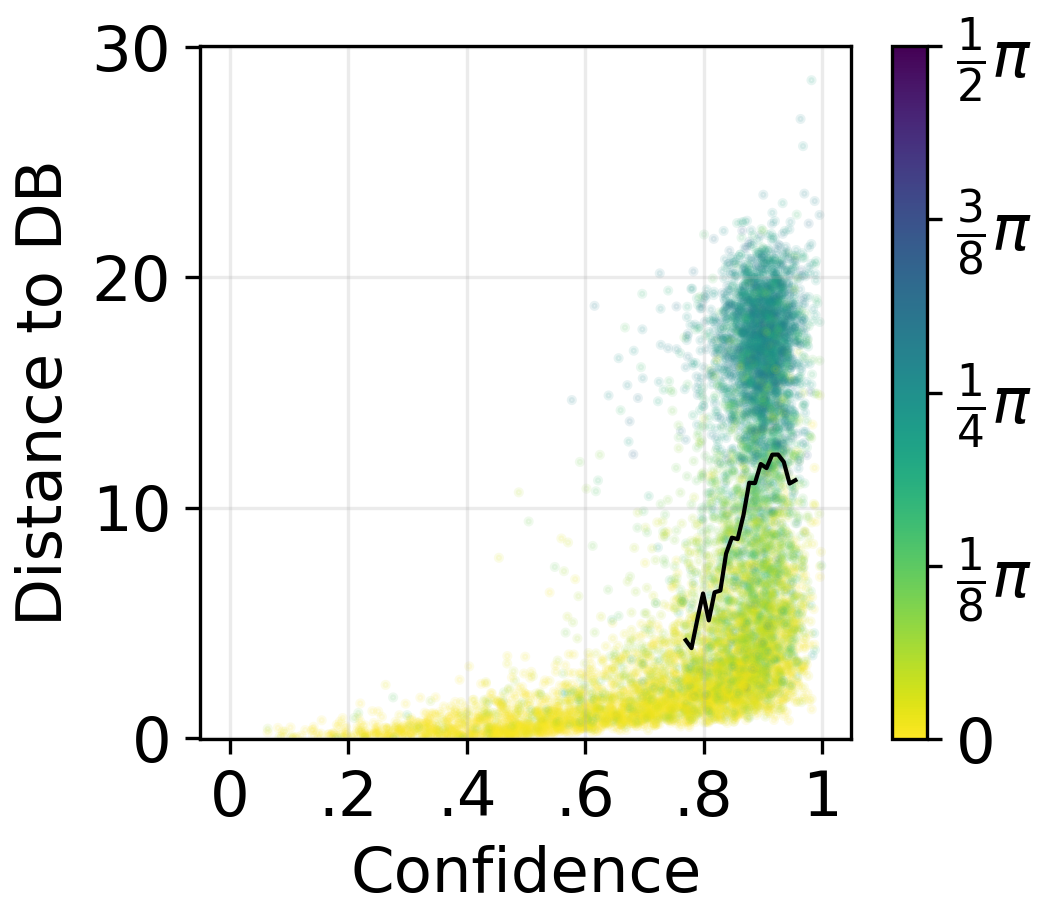} \\
        (a) ViT-B/16 &
        (b) Swin-T
    \end{tabular}

    \caption{
        Distance from the original output to the decision boundary (DB) in the representation space with respect to confidence.
        Colors indicate $\theta^{(1)}$.
        Black lines represent average values.
    }\label{fig:dist_to_db_scatter}
\end{figure}

\subsection{Robustness and underconfidence of Transformers}

Fig.~\ref{fig:dist_to_db_hist} shows the distribution of the Euclidean distance from the original output to the decision boundary in the representation space (i.e., $||\mathbf{z}^{(N)}-\mathbf{z}^{(0)}||$).
Note that the distance scale is different between the models.
Interestingly, the distance distributions for ViT-B/16 and Swin-T are \textit{bimodal}, i.e., the data are grouped into those having small distances and those having large distances.

We examine this phenomenon further in Fig.~\ref{fig:dist_to_db_scatter}, which shows scatter plots between the confidence and the distance, where the colors represent $\theta^{(1)}$ of the corresponding data.
Note that $\theta^{(1)}$ is highly correlated to the total direction change ($\sum_{n=1}^{N-1}{\theta^{(n)}}$), and thus is used as a measure of curvedness of the representation space around the data (see Appendix).
It is clear that the curvedness dichotomizes the data:
those associated with small values of $\theta^{(1)}$ are located in linear regions (marked with yellowish colors), while those associated with large values of $\theta^{(1)}$ are located in curved regions (marked with greenish colors).
In particular, the data in the latter group show larger distances to the decision boundaries, and thus become more robust against adversarial attacks.
In other words, since they are located in curved regions, an attack on them becomes challenging.

\begin{figure}[!t]
    \centering
    \small

    \begin{tabular}{ccccc}
        &
        ResNet50 &
        MobileNetV2 &
        ViT-B/16 &
        Swin-T \\

        \rotatebox[origin=l]{90}{\hspace{7.5mm}$\epsilon_{\mathrm{IFGSM}}=.001$} &
        \includegraphics[width=.20\linewidth]{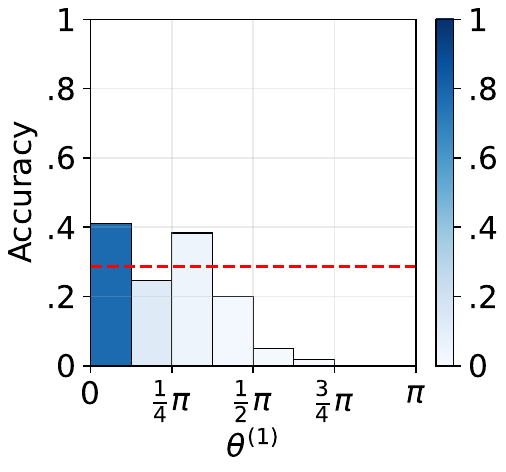} &
        \includegraphics[width=.20\linewidth]{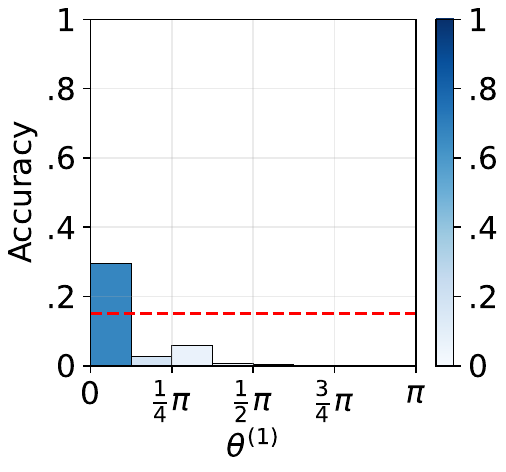} &
        \includegraphics[width=.20\linewidth]{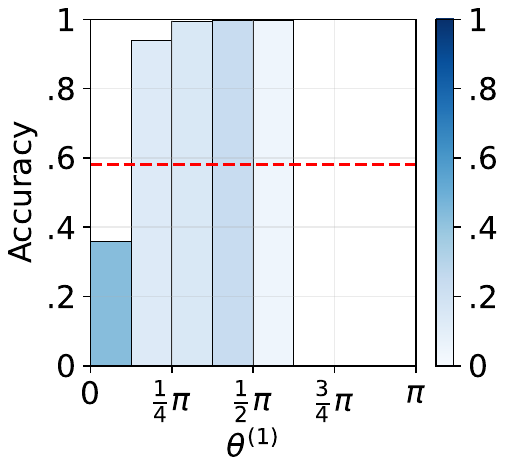} &
        \includegraphics[width=.20\linewidth]{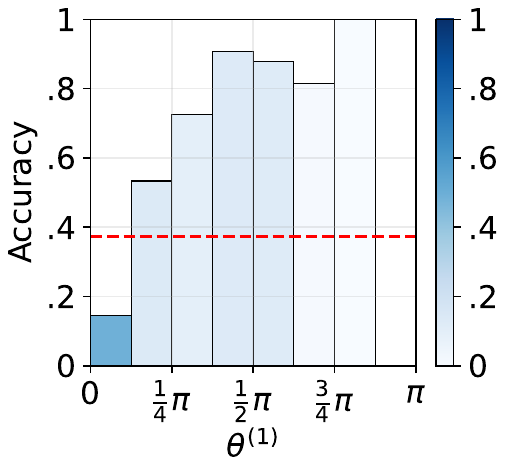} \\
        &
        (a) &
        (b) &
        (c) &
        (d) \\

        \rotatebox[origin=l]{90}{\hspace{7.5mm}$\epsilon_{\mathrm{IFGSM}}=.002$} &
        \includegraphics[width=.20\linewidth]{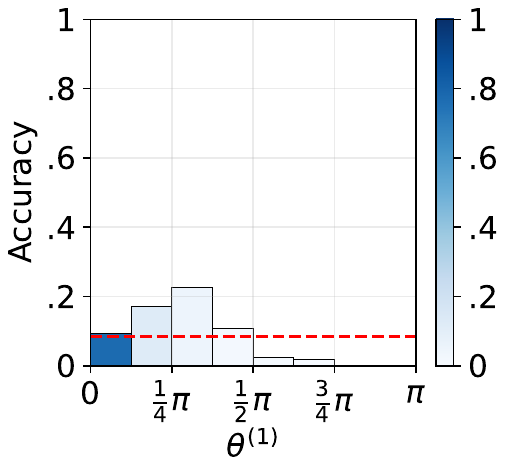} &
        \includegraphics[width=.20\linewidth]{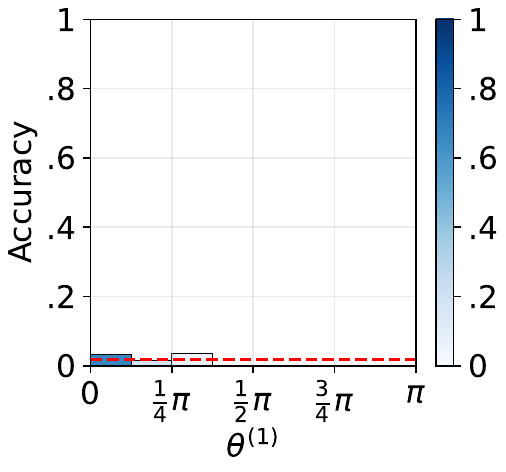} &
        \includegraphics[width=.20\linewidth]{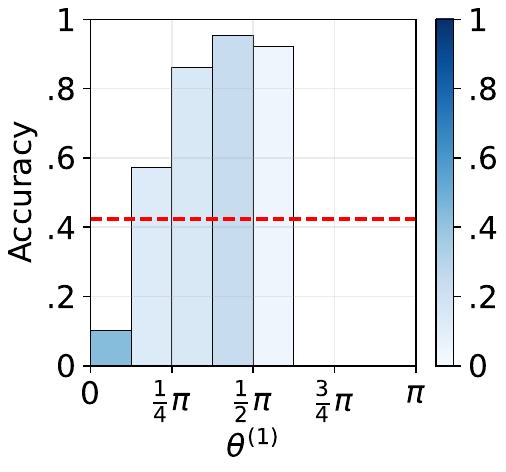} &
        \includegraphics[width=.20\linewidth]{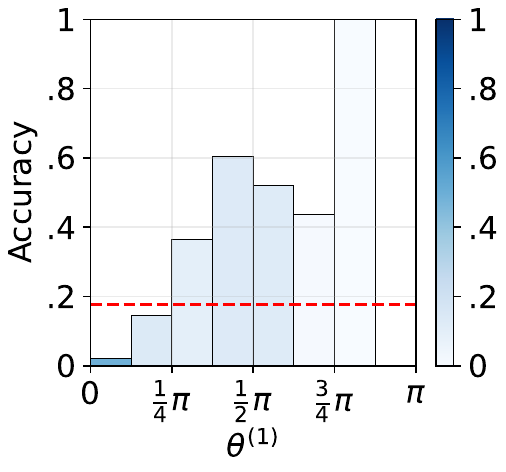} \\
        &
        (e) &
        (f) &
        (g) &
        (h)
    \end{tabular}

    \caption{
        Accuracy after the I-FGSM attack with respect to $\theta^{(1)}$.
        Transparency of the bars represents the ratio of the number of samples in each bin of $\theta^{(1)}$.
        Red dashed lines indicate the overall accuracy after the attack.
    }\label{fig:attack_theta_hist}
\end{figure}

To validate this, we apply the iterative FGSM attack (I-FGSM) \cite{kurakin2017ifgsm}, which is one of the strong attacks, to the correctly classified ImageNet validation data.
We set the maximum amount of perturbation to $\epsilon_{\mathrm{IFGSM}}$=.001 or .002, the number of iterations to $T$=10, and the step size to $\epsilon_{\mathrm{IFGSM}}/T$.
Fig.~\ref{fig:attack_theta_hist} shows the classification accuracy after the attack with respect to $\theta^{(1)}$.
We can observe that the data having large values of $\theta^{(1)}$ show high robustness (i.e., high accuracy even after the attack), which makes the overall robustness of Transformers higher than that of CNNs.

% fig:attack_theta_hist

% fig:effect_of_jump
\begin{figure}[!t]
    \centering
    \small

    \begin{tabular}{cccc}
        \includegraphics[width=.229\linewidth]{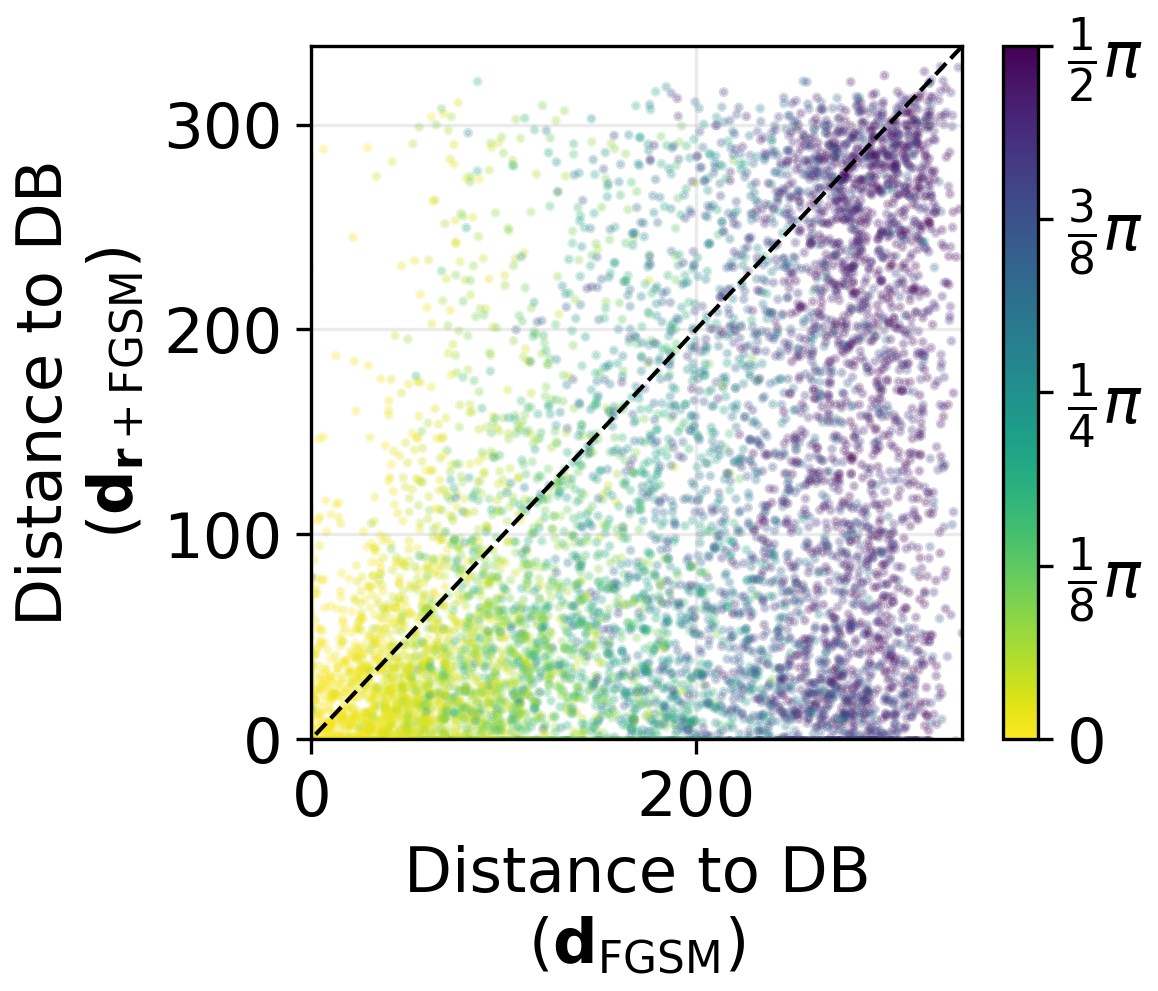} &
        \includegraphics[width=.229\linewidth]{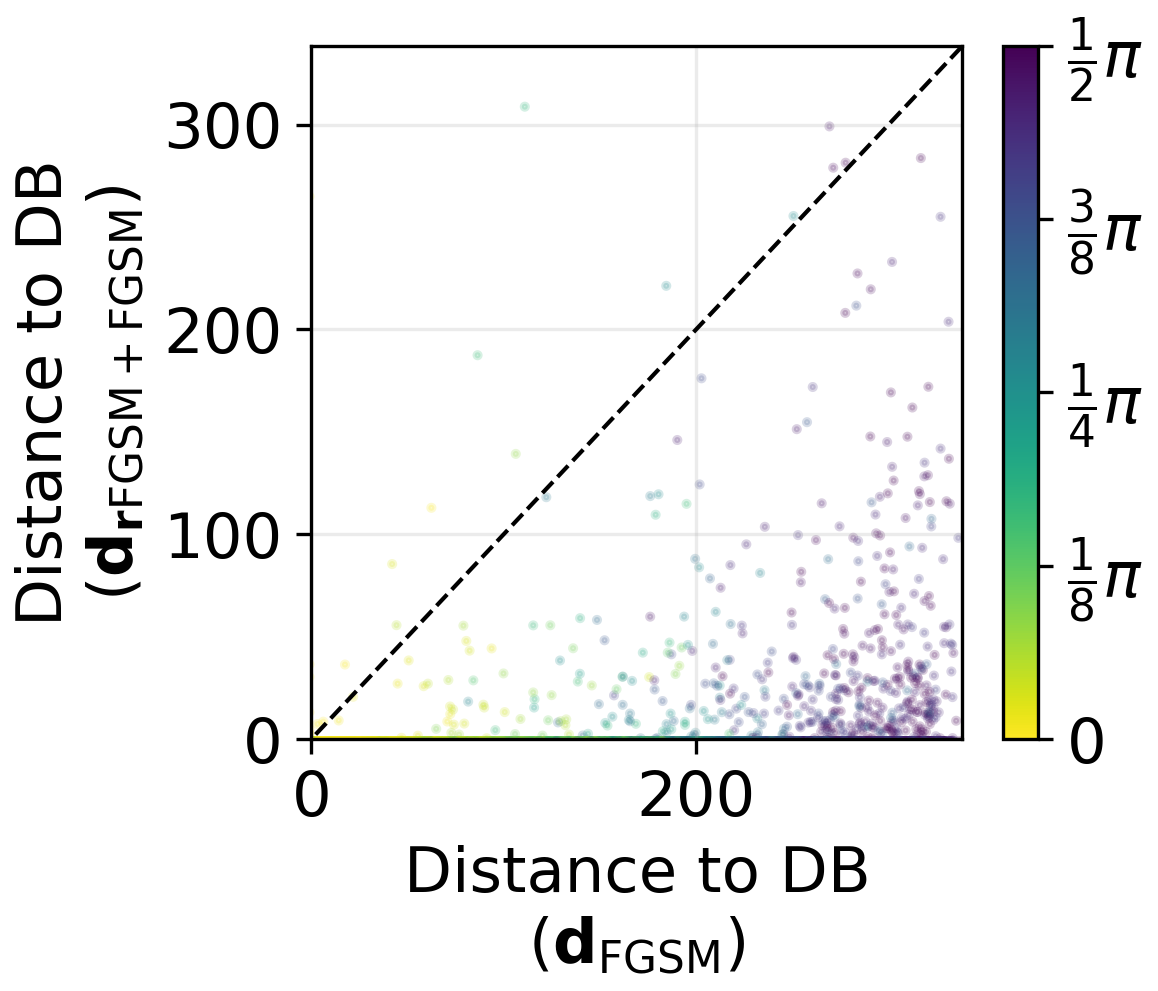} &
        \includegraphics[width=.22\linewidth]{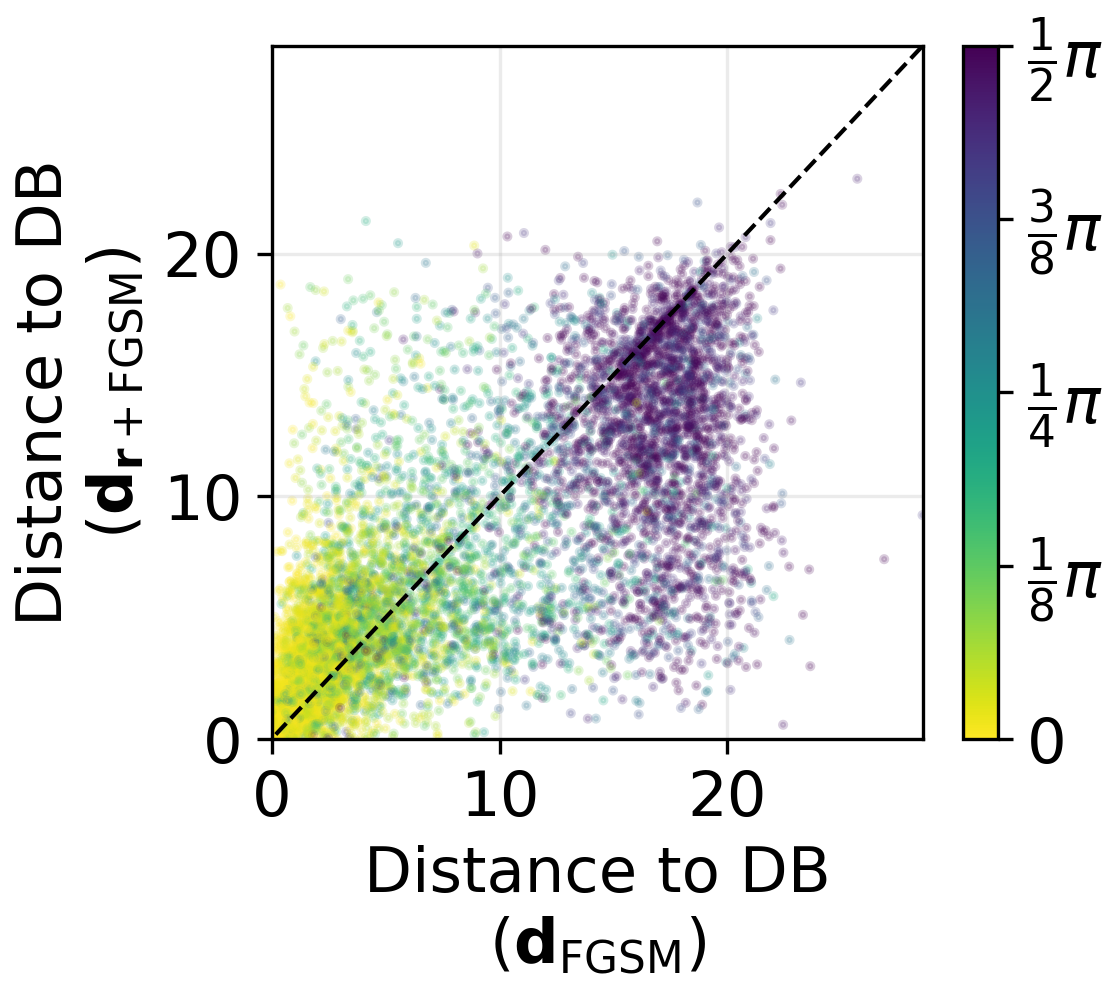} &
        \includegraphics[width=.22\linewidth]{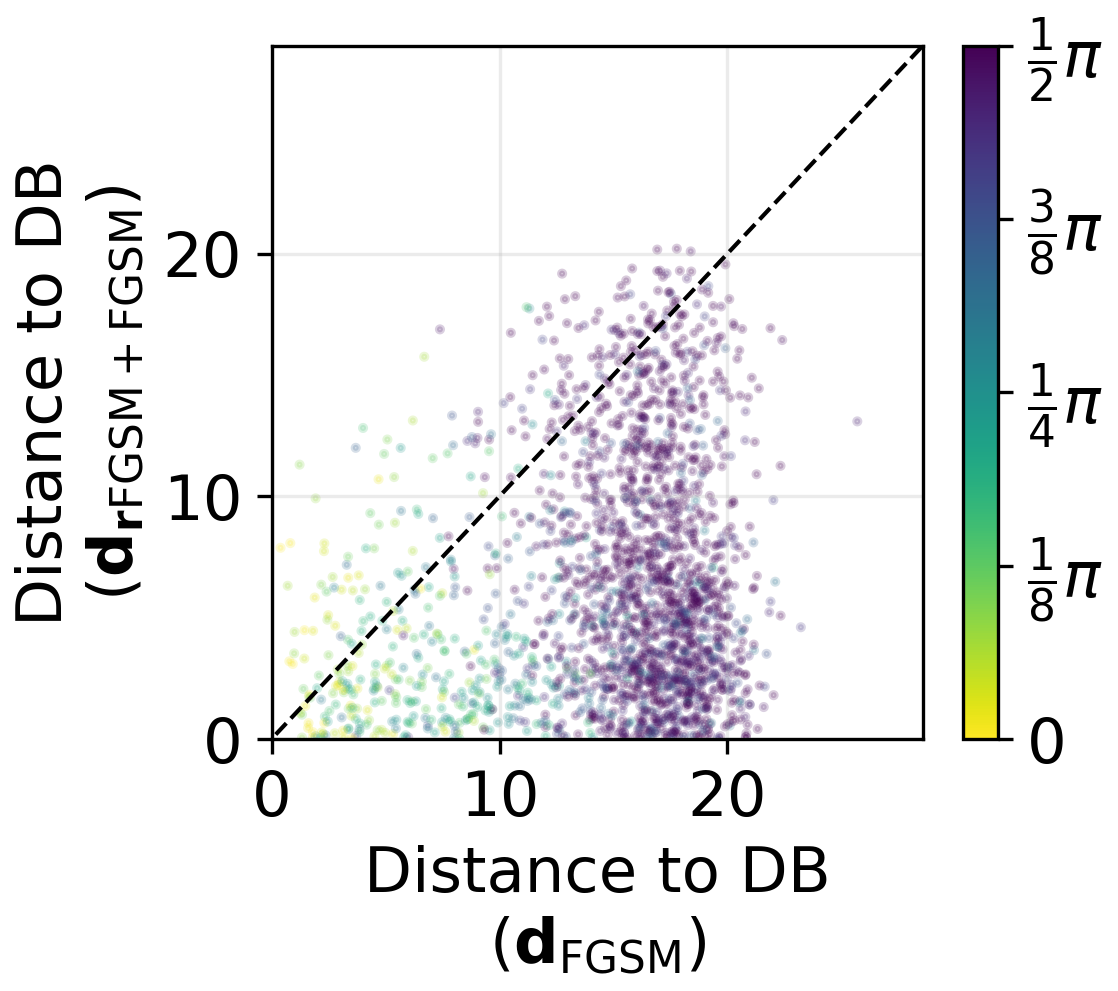} \\
        \multicolumn{2}{c}{(a) ViT-B/16} &
        \multicolumn{2}{c}{(b) Swin-T}
    \end{tabular}

    \caption{
        Relationship of the length of travel ($\epsilon$) to decision boundaries (DB) for original images (x-axis) and jumped images (y-axis).
        Colors indicate $\theta^{(1)}$.
    }\label{fig:effect_of_jump}
\end{figure}

We hypothesize that the curved representation space also causes the underconfident prediction of Transformers.
That is, as shown in Fig.~\ref{fig:illust}b, the decision boundary is actually close to the data point (on the left side of the data), but the curved travel (blue-colored trajectory in Fig.~\ref{fig:illust}b) reaches the decision boundary at a farther location.
To validate this hypothesis, we add a small amount of noise to the input data in order to check if the decision boundary at a closer location can be found if the data jumps out of the curved region (i.e., reaching $\mathbf{z_r}^{(N)}$ from $\mathbf{z_r}^{(0)}$ in Fig.~\ref{fig:illust}b).

Figs.~\ref{fig:effect_of_jump}a and \ref{fig:effect_of_jump}b show the relationship of the distance to decision boundaries for original outputs (x-axis) and jumped images (y-axis) for Swin-T.
The direction for travel is indicated in the axis.
It can be observed that when the FGSM direction is computed and used after random jump ($\mathrm{\mathbf{d}} = \mathrm{\mathbf{d}}_{\mathrm{\mathbf{r}} + \mathrm{FGSM}}$; yellow-colored trajectory in Fig.~\ref{fig:illust}), the distance is significantly reduced (left figures in Figs.~\ref{fig:effect_of_jump}a and \ref{fig:effect_of_jump}b; most data points under the 45$^\circ$ line).
As shown in Figs.~\ref{fig:footprint_csim}g and \ref{fig:footprint_csim}h, the travel becomes less curved and thus the decision boundary can be reached effectively.
The 2D projected movements after the jump in Fig.~\ref{fig:visualize_jump} in Appendix also supports this.
Furthermore, the random jump can be made even more effective by setting the jump direction to the FGSM direction that would have been found if random jump was applied ($\mathrm{\mathbf{d}} = \mathrm{\mathbf{d}}_{\mathrm{\mathbf{r}FGSM} + \mathrm{FGSM}}$; red-colored trajectory in Fig.~\ref{fig:illust}), resulting in further reduction in distance (right figures in Figs.~\ref{fig:effect_of_jump}a and \ref{fig:effect_of_jump}b).

% fig:attacked_samples
\begin{figure}[!t]
    \centering
    \small

    \includegraphics[width=.9\linewidth]{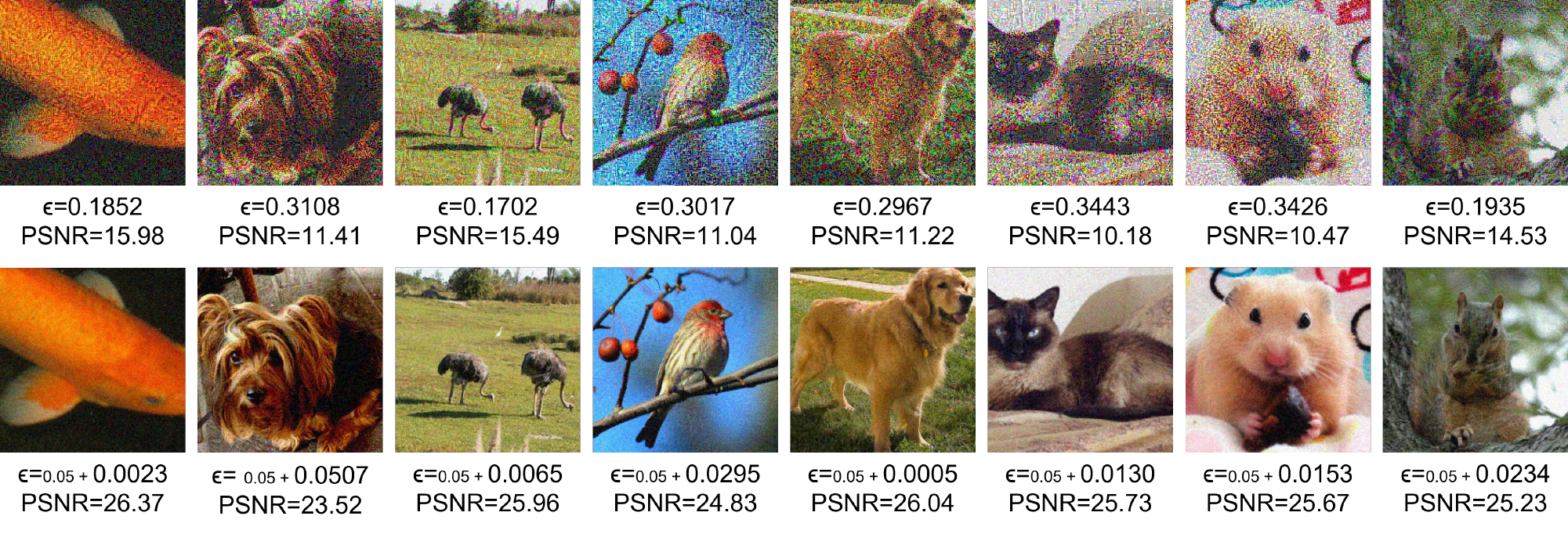}

    \caption{
        Example images that are perturbed by FGSM so as to reach decision boundaries and become misclassified.
        The total amount of perturbation ($\epsilon$) and the peak signal-to-noise ratio (PSNR) in dB are also shown.
        \textbf{Top}: Perturbed images.
        \textbf{Bottom}: Images perturbed after random jump ($\epsilon_{\mathrm{\mathbf{r}}}=.05$).
    }\label{fig:attacked_samples}
\end{figure}

The reduced distance to the boundary by random jump implies that the jumped input data can be made misclassified by adding a smaller amount of perturbation than the original input data.
Fig.~\ref{fig:attacked_samples} demonstrates that this actually works. The figure shows example images perturbed linearly in the FGSM direction (i.e., $\mathbf{x}^{(N)}$) and those first undergone random jump ($\epsilon_\mathbf{r}$=.05) and then perturbed linearly in the FGSM direction (i.e., $\mathbf{x_r}^{(N)}$) for Swin-T. It is clear that the images are easily misclassified with significantly reduced amounts of perturbation (smaller $\epsilon$ and higher PSNR) after the random jump passing over curved regions.

\subsection{Nonlinearity of attention operation}

Why do curves tend to appear in the representation space of Transformers only, and not in CNNs?
In this section, we explain this theoretically by revisiting convolution and self-attention operations.
Note that we use matrices to denote inputs and outputs instead of vectors for better explanation of the operations.
Empirical results of this section can be found in the next section and Table~\ref{tab:xray} in Appendix.

Deep neural networks transform data points through contiguous blocks that perform similar operations.
CNNs, for instance, comprise layers of a convolution operation and activation.
As well known, a convolution is a linear operation \cite{hayes1996dsp}, i.e., an increment $\mathrm{\mathbf{P}}$ to the input $\mathrm{\mathbf{X}}$ converts into the addition of separate responses:
\begin{equation}
    \textrm{Conv}(\mathrm{\textbf{X}} + \mathrm{\mathbf{P}}) =
    \textrm{Conv}(\mathrm{\textbf{X}}) + \mathrm{Conv}(\mathrm{\mathbf{P}}).
    \label{eq:conv}
\end{equation}
Activation functions may imbue the transformation with nonlinearity in theory, which is very limited in reality.
ReLUs are linear until the input data travels to the negative region.
In the case of sigmoid functions, the input data is supposed to linger in the non-saturated region, which is pseudo-linear.
Therefore, the main building block of CNNs is a linear transformation.

At the heart of Transformers, an attention block transforms an input \textit{\textbf{query}} into the weighted sum of neighbor \textit{\textbf{values}}, which is a linear projection of input tokens.
The weights are calculated as softmax of attention scores $\mathrm{\mathbf{A}}$, which is an inner product of \textit{\textbf{query}} and \textit{\textbf{key}}:
\begin{align}
    \mathrm{\mathbf{A}}(\mathrm{\mathbf{X}}) &=
    \mathrm{\mathbf{X}}\mathrm{\mathbf{W_{q}}}
    \mathrm{\mathbf{W^{\top}_{k}}}\mathrm{\mathbf{X}^{\top}},
    \label{eq:attn_scores} \\
    \textrm{Attn}(\mathrm{\mathbf{X}}) &=
    \textrm{softmax}(\mathrm{\mathbf{A}}/\sqrt{D_{k}})\mathrm{\mathbf{X}}\mathrm{\mathbf{W_{v}}},
    \label{eq:attn}
\end{align}
where $\mathrm{\mathbf{W_{q}}}$, $\mathrm{\mathbf{W_{k}}}$, and $\mathrm{\mathbf{W_{v}}}$ are the projection heads for query, key, and value, respectively, and $D_{k}$ is the column dimension of $\mathrm{\mathbf{W_{k}}}$.
If $\mathrm{\mathbf{X}}$ is moved by $\mathrm{\mathbf{P}}$, $\mathrm{\mathbf{A}}$ will change as follows:
\begin{align}
    \mathrm{\mathbf{A}}(\mathrm{\mathbf{X}} + \mathrm{\mathbf{P}}) = &
    (\mathrm{\mathbf{X}} + \mathrm{\mathbf{P}})
    \mathrm{\mathbf{W_{q}}}
    \mathrm{\mathbf{W^{\top}_{k}}}
    (\mathrm{\mathbf{X}}^{\top} + \mathrm{\mathbf{P}}^{\top})
    \label{eq:attn_moved_1} \\
    = &
    \mathrm{\mathbf{X}}\mathrm{\mathbf{W_{q}}}\mathrm{\mathbf{W^{\top}_{k}}}\mathrm{\mathbf{X}^{\top}} +
    \mathrm{\mathbf{P}}\mathrm{\mathbf{W_{q}}}\mathrm{\mathbf{W^{\top}_{k}}}\mathrm{\mathbf{P}^{\top}} + \nonumber \\ &
    \mathrm{\mathbf{X}}\mathrm{\mathbf{W_{q}}}\mathrm{\mathbf{W^{\top}_{k}}}\mathrm{\mathbf{P}^{\top}} +
    ~\mathrm{\mathbf{P}}\mathrm{\mathbf{W_{q}}}\mathrm{\mathbf{W^{\top}_{k}}}\mathrm{\mathbf{X}^{\top}}
    \label{eq:attn_moved_3} \\
    = &
    \mathrm{\mathbf{A}}(\mathrm{\mathbf{X}}) +
    \mathrm{\mathbf{A}}(\mathrm{\mathbf{P}}) + \nonumber \\ &
    \underbrace{
    \mathrm{\mathbf{X}}\mathrm{\mathbf{W_{q}}}\mathrm{\mathbf{W^{\top}_{k}}}\mathrm{\mathbf{P}^{\top}} +
    \mathrm{\mathbf{P}}\mathrm{\mathbf{W_{q}}}\mathrm{\mathbf{W^{\top}_{k}}}\mathrm{\mathbf{X}^{\top}}
    }_\text{residual}.
    \label{eq:attn_moved_residual}
\end{align}
As shown in Eq.~\ref{eq:attn_moved_residual}, the attention score is not linear and the deviation from the linear response is the combination of the projection heads and input data.
During the inference operation, the projection heads are fixed and the linear perturbation to the input data will generate a varying degree of nonlinearity depending on the magnitude of the input and the angle between the projection head and the input (see Fig.~\ref{fig:attn_data_dependent} in Appendix for detailed discussion).
Additionally, the softmax function in Eq.~\ref{eq:attn} augments the nonlinearity of the attention operation.

\section{Additional intriguing observations}

In this section, in addition to the aforementioned main discoveries, we share our additional intriguing observations, for which we leave further detailed analysis as future work.

\subsection{Contribution of components to curvedness}

Which component in Transformers fortifies the curvedness of the representation space?
When ResNet50 and Swin-T are compared (Table~\ref{tab:xray} in Appendix), we find that in both models the activation functions contribute the most to the increase of $\theta^{(1)}$.
GELU causes curvedness more than ReLU because the former is more nonlinear than the latter.
In the case of ResNet50, the convolutional layers and batch normalization (BatchNorm) do not cause curvedness of the representation space.
In contrast, for Swin-T, the layer normalization (LayerNorm) and self-attention layers intensify the curvedness.
The result of these compound contributions of different components appears as the curvedness of the representation space of Transformers.

This observation raises an interesting question about
ConvNeXt~\cite{liu2022convnext}, which is a CNN but uses GELU and LayerNorm instead of ReLU and BatchNorm, respectively:
Which trend will it follow, CNNs or Transformers?
Surprisingly, we observe that ConvNeXt-Tiny follows the trend of \textit{Transformers}, rather than CNNs (Table~\ref{tab:xray} and Figs.~\ref{fig:calibration_3}a and \ref{fig:fgsm-travel_3}a in Appendix).
This indicates that the curvedness in the representation space highly depends on the particular components used in models, and is not just a problem of models being CNNs or Transformers.

\subsection{Knowledge distillation and curvedness}

Another interesting model we find is DeiT-Ti~\cite{touvron2021deit}, a convolution-free Transformer, and its distilled version, which we refer to as DeiT-Ti-Distilled.
We observe that as expected, DeiT-Ti follows the trend of Transformers, i.e., underconfidence with high robustness (Figs.~\ref{fig:calibration_3}b and \ref{fig:fgsm-travel_3}b in Appendix).
However, DeiT-Ti-Distilled, knowledge-distilled DeiT-Ti with a CNN teacher, tends to follow the trend of CNNs, i.e., overconfidence with low robustness (Figs.~\ref{fig:calibration_3}c and \ref{fig:fgsm-travel_3}c in Appendix).
The results in Table~\ref{tab:xray} also coincides with this observation, where the values of $\theta^{(1)}$ are reduced for DeiT-Ti-Distilled compared to DeiT-Ti.
This indicates that knowledge distillation can also affect the nonlinearity of Transformers.

\subsection{Curved space during training}

For deeper understanding of the curved regions in the representation space, we look into the training process of Transformers.
We observe that for the data located in curved regions, the loss does not change much from the early training stage (Fig.~\ref{fig:train_loss_change} in Appendix; no change in loss for bottom rows in the figure, which show large values of $\theta^{(1)}$).
This phenomenon can also be observed from another view, in terms of the relationship between the loss at a certain epoch and the loss change from the epoch until the end of training (Fig.~\ref{fig:train_loss-loss} in Appendix; loss values for the data residing in curved regions - dark-colored points in the figure - are hardly reduced already from 30 epochs).
In other words, certain training data seem to be \textit{trapped} in curved regions, which obstructs the training of the network.

When do curved regions start to form?
When we check the relationship of $\theta^{(1)}$ at a certain training stage and $\theta^{(1)}$ after training, we observe that once a data is trapped in a curved region, it hardly escapes the region and $\theta^{(1)}$ becomes larger during training, i.e., the curvedness gets severer (Fig.~\ref{fig:train_theta-theta} in Appendix; data points mostly above the 45$^\circ$ line).

\section{Conclusion}

We studied the input-output relationship of Transformers by examining the trajectory of the output in the representation space with respect to linear movements in the input space.
The experimental results indicated that the representation space of Transformers is curved around some data, which explains high robustness and underconfident prediction of Transformers.

\section{Future work}

In general, understanding the behavior of a certain neural network model, either analytically or empirically, is a difficult task, which cannot be accomplished by a single paper but requires a lot of research efforts.
We have focused on the input-output relationship along the adversarial direction generated by FGSM, which revealed the existence of curvedness in the representation space of Transformers.
We believe that we have opened a new perspective of understanding Transformers, and many derivative research questions will naturally follow, e.g., consideration of different travel directions, input-output relationship of various building blocks of neural networks, effects of different training recipes, effects of training datasets, etc., which we leave as future works.
It is also our hope that our work promotes further interesting research topics (e.g., ways to reduce/intensify curvedness during or after training, measures of local/global curvedness, theoretical analysis of curvedness, etc.) and applications (e.g., effective adversarial attacks considering curvedness, robust model architectures, robust training methods, etc.).

%Bibliography
\bibliographystyle{unsrt}  
\bibliography{ref}

%%%%%%%%%%%%
% APPENDIX %
%%%%%%%%%%%%

\clearpage
\appendix

\section{The common intuition of robustness and confidence}

A data on the decision boundary between two classes has the equal confidence values for the two classes (e.g., 0.5 for binary classification).
When we gradually move the data from the boundary to the decision region of one class, the confidence for the class will gradually increase for a certain range.
For traditional machine learning classifiers, the positive correlation between the distance to the decision boundary and the confidence is easily expected (e.g., linear classifier, as mentioned in Section 6.8 of~\cite{watt2020mlbook}).
This trend is also shown for deep models: for deep fully connected networks~\cite{yousefzadeh2019investigating}, for LeNet~\cite{pearce2021understanding}, and for CNNs~\cite{hein2019relu}.
Moreover, the trend is used for efficient classification using CNNs~\cite{panda2016conditional}, robustness analysis of CNNs~\cite{yang2020boundary}, and performance analysis of CNNs~\cite{hu2023aries}.
However, for Transformers, the validity of the trend has not been examined before, and we find that the trend does not hold for Transformers.

\section{Representation space visualization}

Figs.~\ref{fig:visualize} and \ref{fig:visualize_jump} are produced as follows.
The FGSM direction $\mathrm{\mathbf{d}}_{\mathrm{\mathbf{x}}}$ is found for the given data $\mathrm{\mathbf{x}}$, and we choose a random direction $\mathrm{\mathbf{d}}_{\mathrm{\mathbf{x}}}^{\perp}$ orthogonal to $\mathrm{\mathbf{d}}_{\mathrm{\mathbf{x}}}$.
Then, we move $\mathrm{\mathbf{x}}$ in the input space using $\mathrm{\mathbf{d}}_{\mathrm{\mathbf{x}}}$ and $\mathrm{\mathbf{d}}_{\mathrm{\mathbf{x}}}^{\perp}$ by
\begin{equation}
    \mathrm{\mathbf{x}}_{ij} =
        \mathrm{\mathbf{x}} +
        \frac{\alpha \cdot i}{M} \mathrm{\mathbf{d}}_{\mathrm{\mathbf{x}}} +
        \frac{\alpha \cdot j}{M} \mathrm{\mathbf{d}}_{\mathrm{\mathbf{x}}}^{\perp}
\end{equation}
where $\alpha$ is a positive real number, $M$ determines the total number of points in the grid of the input space as $(2M+1)^{2}$ (Fig.~\ref{fig:visualize}a), and $i$ and $j$ are integers within $[-M, M]$.

For each $\mathrm{\mathbf{x}}_{ij}$, we obtain its output at the penultimate layer of a model, $\mathrm{\mathbf{z}}_{ij}$, i.e., the feature in the representation space.
Then, $\mathrm{\mathbf{z}}_{ij}$ is projected onto the two-dimensional plane determined by two arbitrary mutually orthogonal vectors $\mathrm{\mathbf{d}}_{\mathrm{\mathbf{z}}}$ and $\mathrm{\mathbf{d}}_{\mathrm{\mathbf{z}}}^{\perp}$, which produce Figs.~\ref{fig:visualize}b, \ref{fig:visualize}c, and \ref{fig:visualize_jump}.

\section{Model calibration}

\begin{figure}[!t]
    \centering
    \small

    \begin{tabular}{cccc}
        \includegraphics[width=.20\linewidth]{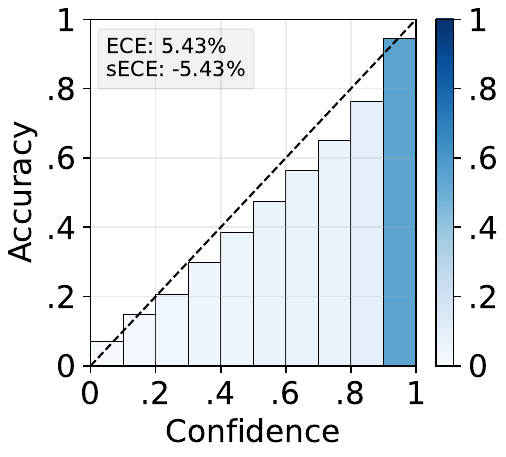} &
        \includegraphics[width=.20\linewidth]{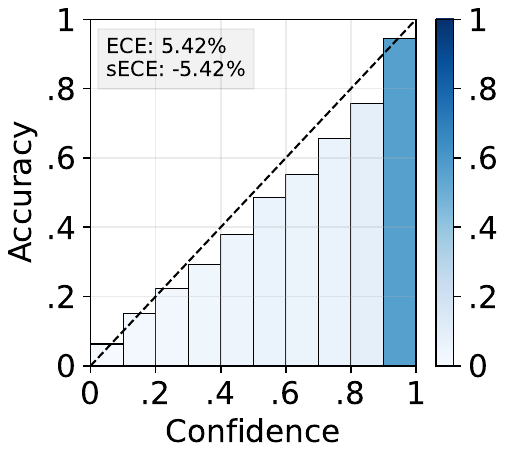} &
        \includegraphics[width=.20\linewidth]{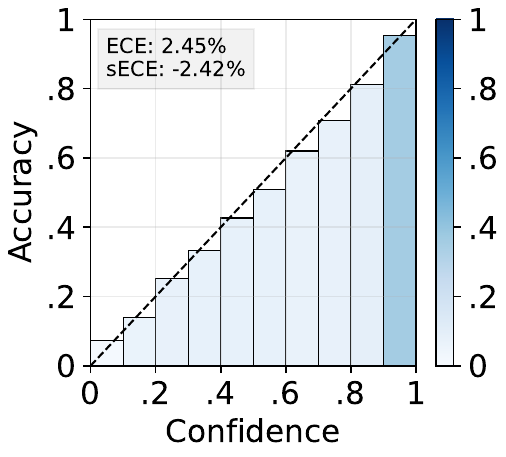} &
        \includegraphics[width=.20\linewidth]{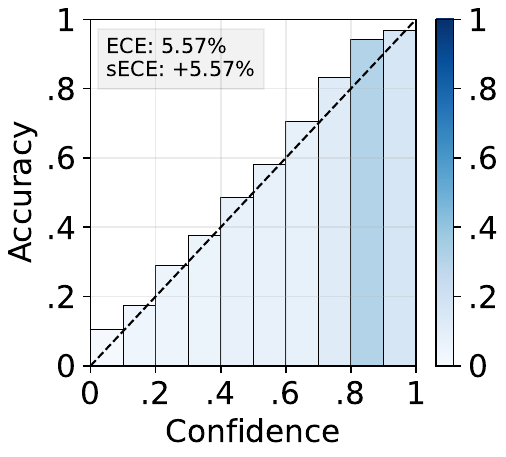} \\
        (a) ResNet101 &
        (b) ResNet152 &
        (c) MobileNetV3-Small &
        (d) ViT-B/32 \\

        \includegraphics[width=.20\linewidth]{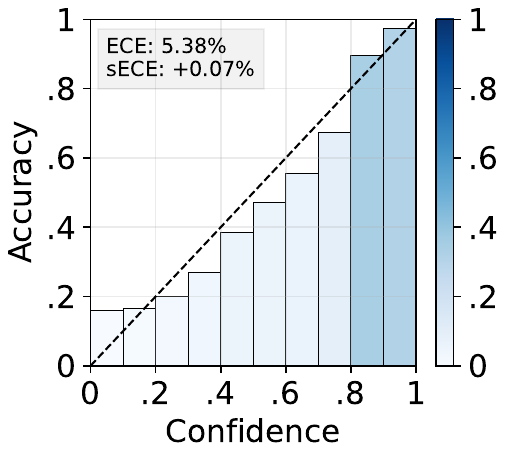} &
        \includegraphics[width=.20\linewidth]{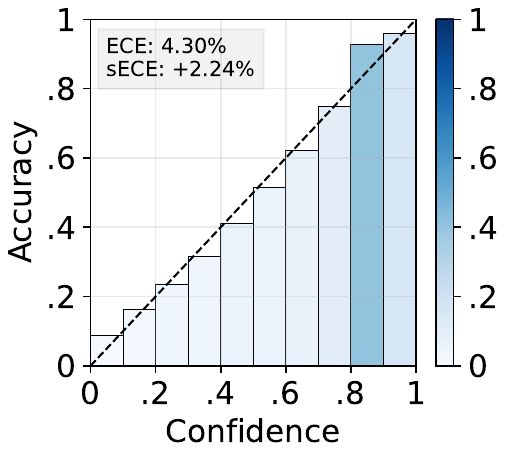} &
        \includegraphics[width=.20\linewidth]{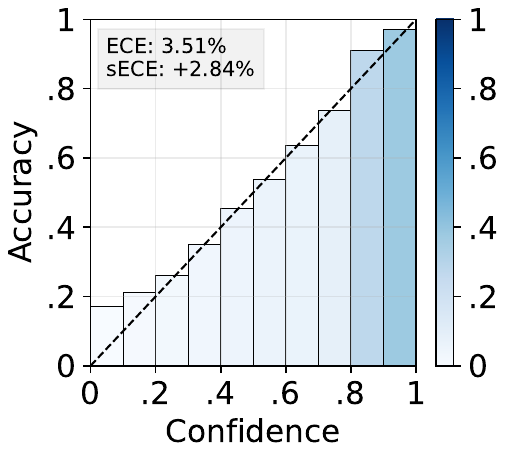} &
        \includegraphics[width=.20\linewidth]{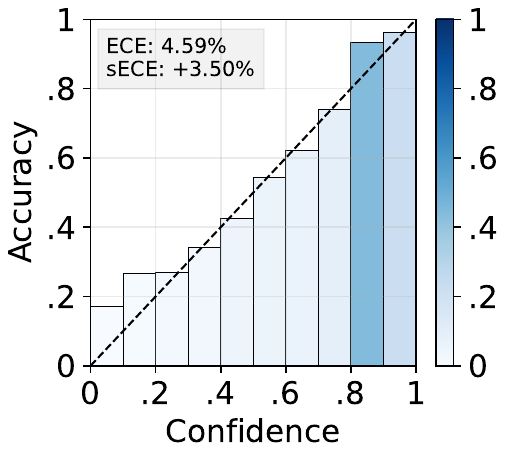} \\
        (e) ViT-L/16 &
        (f) ViT-L/32 &
        (g) Swin-S &
        (h) Swin-B
    \end{tabular}

    \caption{
        Reliability diagrams of CNNs and Transformers.
        Transparency of bars represent the ratio of the number of data in each confidence bin.
        ECE and sECE values are also shown in each case.
    }\label{fig:calibration_2}
\end{figure}

Fig.~\ref{fig:calibration_2} extends Fig.~\ref{fig:calibration} for models with different sizes or versions.
It can be observed that the calibration trends for CNNs (negative sECE, i.e., overconfident) and Transformers (positive sECE, i.e., underconfident) are consistent.

\section{Passage to decision boundary}

Algorithm~\ref{alg:travel} provides the detailed procedure to travel to the decision boundary linearly in the input space.

\begin{figure}[!t]
    \centering
    \small

    \begin{tabular}{cccc}
        \includegraphics[width=.20\linewidth]{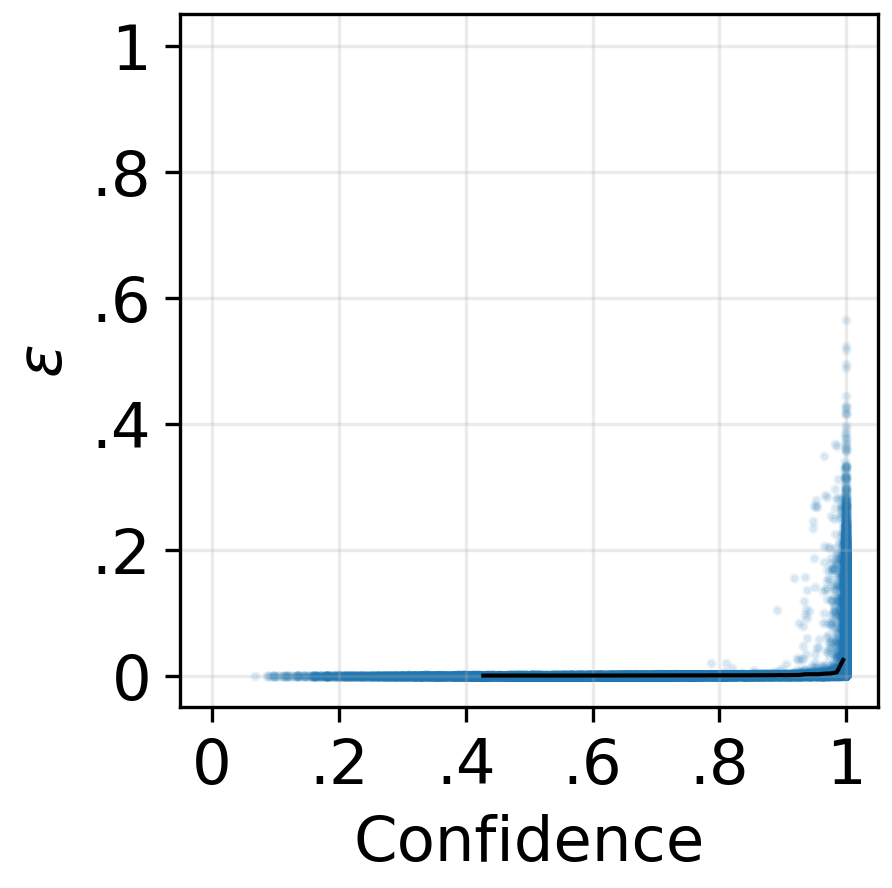} &
        \includegraphics[width=.20\linewidth]{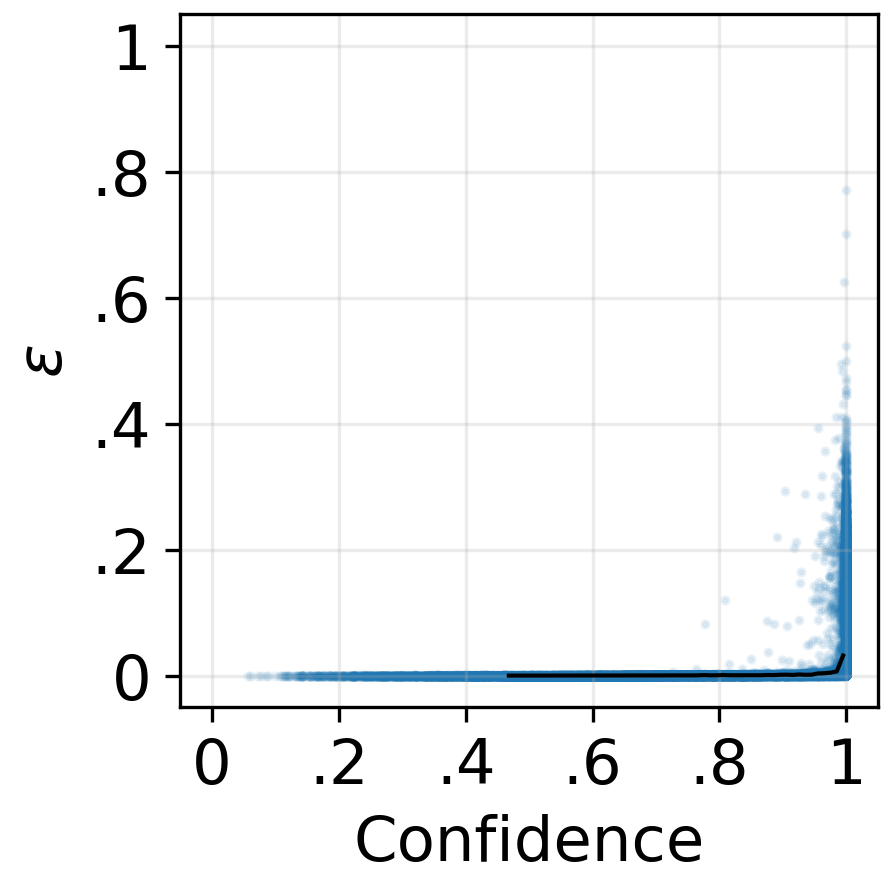} &
        \includegraphics[width=.20\linewidth]{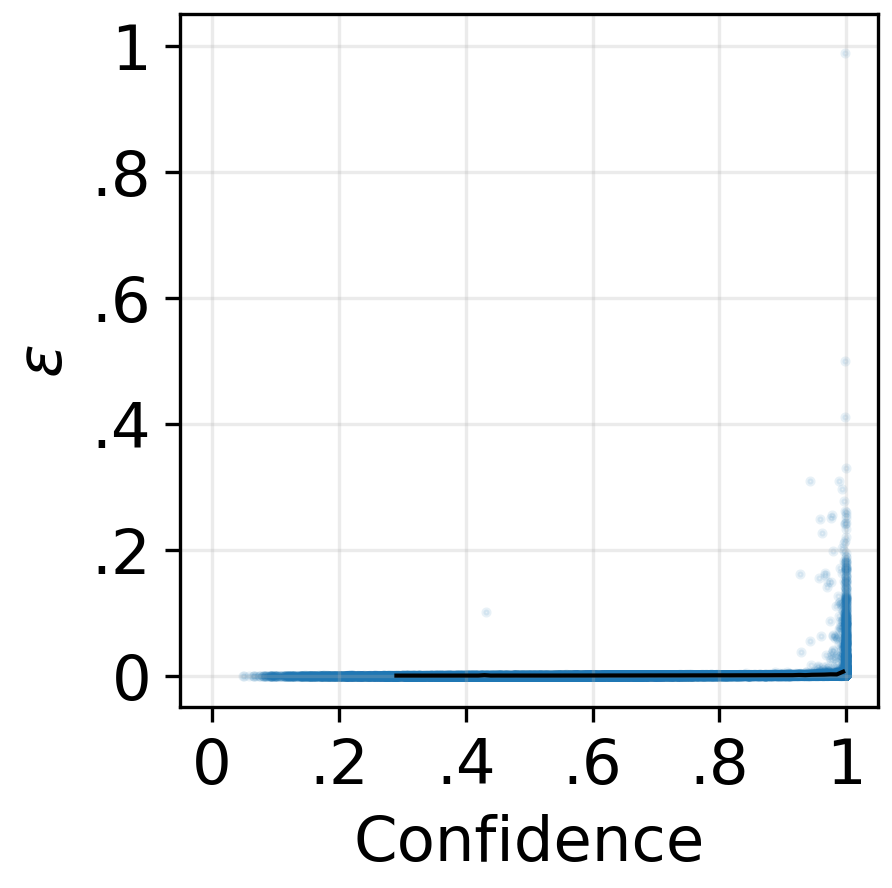} &
        \includegraphics[width=.20\linewidth]{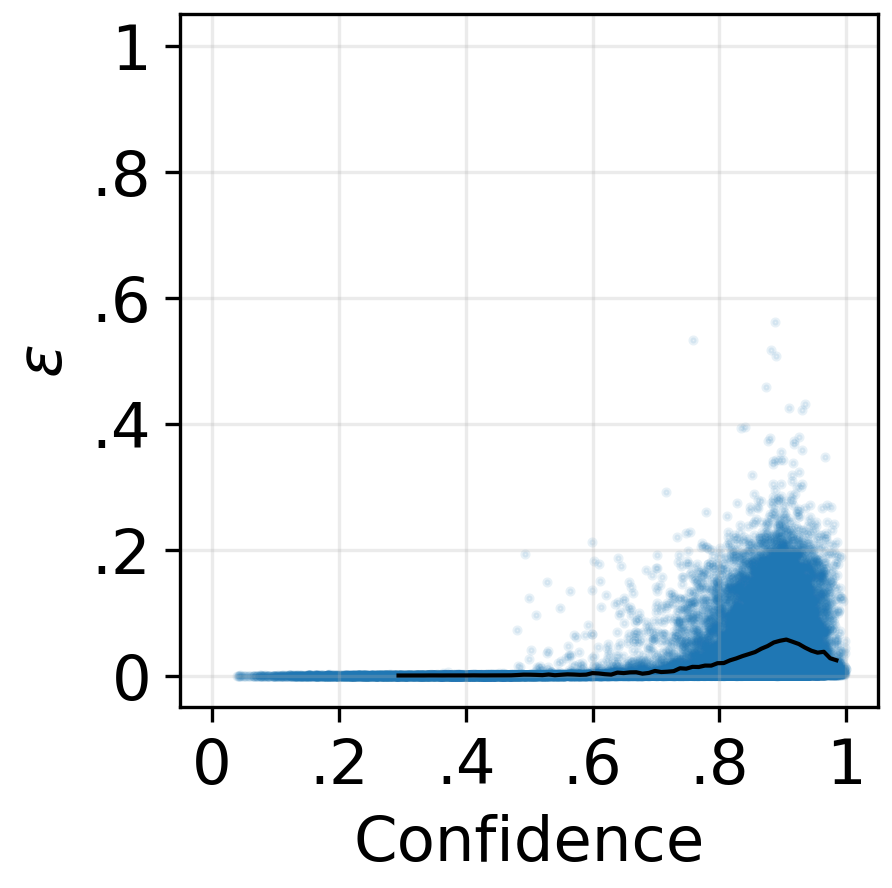} \\
        \hspace{5.0mm} (a) ResNet101 &
        \hspace{5.0mm} (b) ResNet152 &
        \hspace{5.0mm} (c) MobileNetV3-Small &
        \hspace{5.0mm} (d) ViT-B/32 \\
        \includegraphics[width=.20\linewidth]{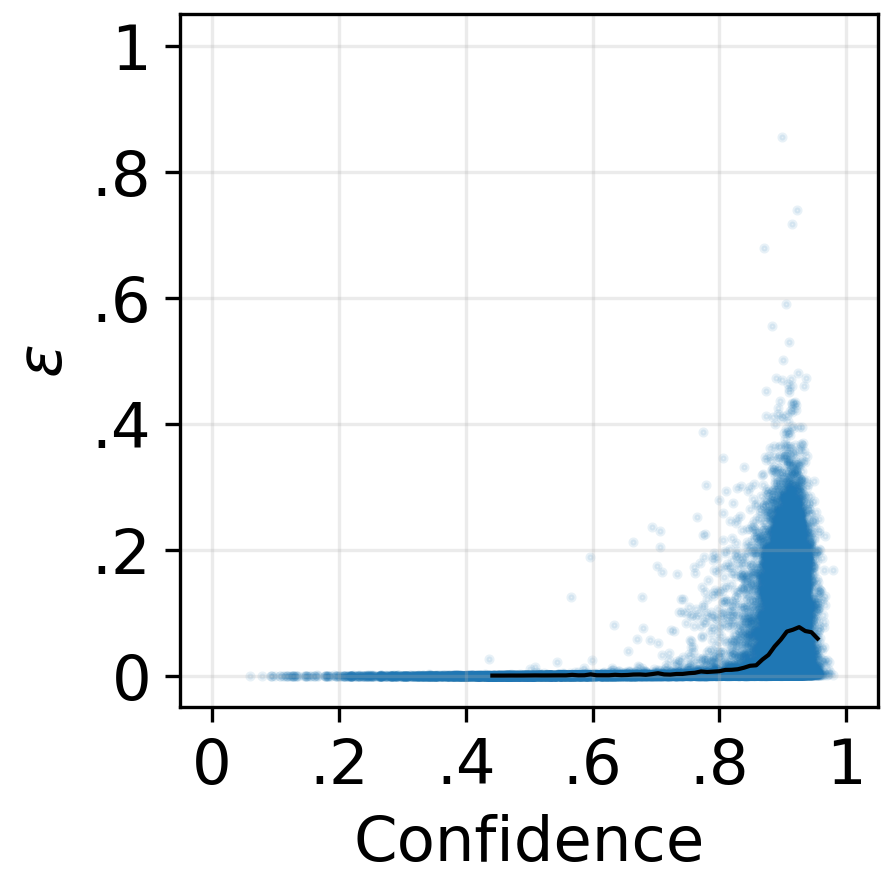} &
        \includegraphics[width=.20\linewidth]{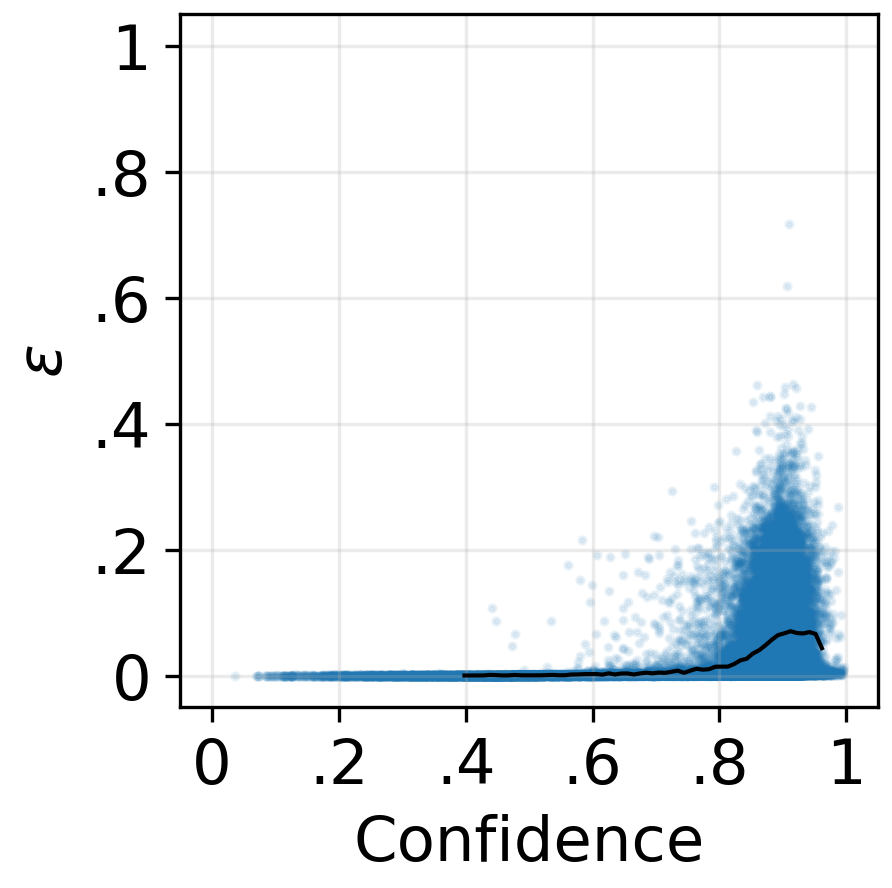} &
        \includegraphics[width=.20\linewidth]{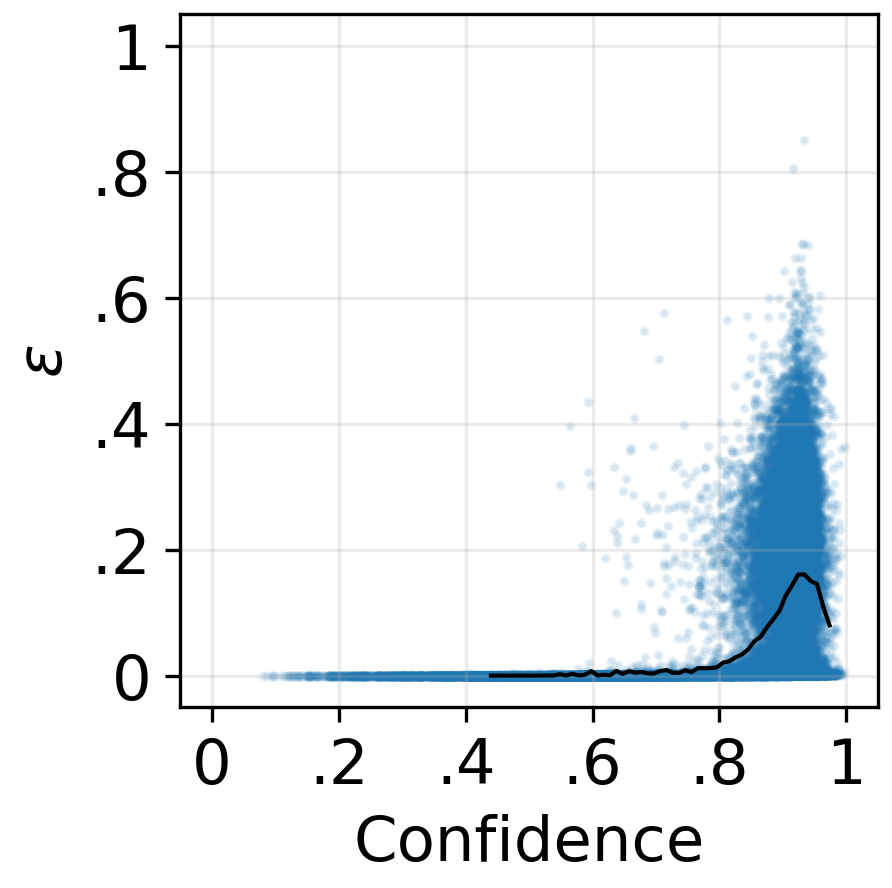} &
        \includegraphics[width=.20\linewidth]{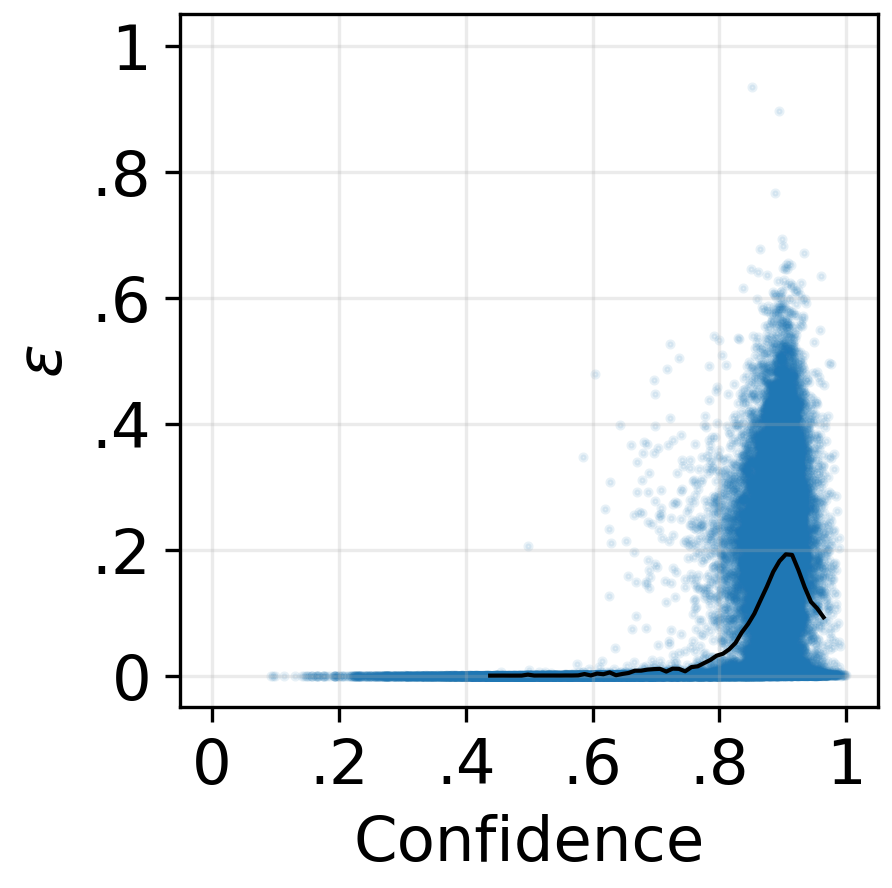} \\
        \hspace{5.0mm} (e) ViT-L/16 &
        \hspace{5.0mm} (f) ViT-L/32 &
        \hspace{5.0mm} (g) Swin-S &
        \hspace{5.0mm} (h) Swin-B
    \end{tabular}

    \caption{
        Lengths ($\epsilon$) of the travel to decision boundaries with respect to the confidence for the ImageNet validation data.
        Black lines represent average values.
    }\label{fig:fgsm-travel_2}
\end{figure}

Fig.~\ref{fig:fgsm-travel_2} extends Fig.~\ref{fig:fgsm-travel} for models with different sizes or versions.

\begin{algorithm}[!ht]
\caption{Travel to decision boundary}

\textbf{Input: }A correctly classified image $\mathrm{\mathbf{x}}\in[0, 1]^{D}$, true class label $y$, classifier $\mathcal{C}$, travel direction $\mathrm{\mathbf{d}}$ \\
\textbf{Parameters: } Initial length of travel $\epsilon_i$, travel length $\epsilon$, length decay $\epsilon_d$ ($0<\epsilon_d<1$), tolerance $\epsilon_t$ \\
\textbf{Output: }Traveled image $\mathrm{\mathbf{x}}^\prime$ \\

% alg:travel
\begin{algorithmic}[1]
\vspace{-4mm}

\STATE $\epsilon \gets \epsilon_{i}$
\STATE Define travel stride $\Delta \epsilon \gets \epsilon_{i}$.

\WHILE{$\epsilon < \epsilon_t$}

    \STATE Generate perturbation $\mathrm{\mathbf{p}}=\epsilon \cdot \mathrm{\mathbf{d}}$.
    
    \STATE $\mathrm{\mathbf{x}}^\prime \gets \mathrm{clip}_{0,1}(\mathrm{\mathbf{x}}+\mathrm{\mathbf{p}})$
    \item where $\mathrm{clip}_{a,b}(z) = \min(\max(z,a),b)$.
    
    \IF{$\mathcal{C}(\mathrm{\mathbf{x}}^\prime) \neq \mathrm{\mathbf{y}}$}
        \item $\epsilon \gets \epsilon - \Delta \epsilon$
        \item $\Delta \epsilon \gets \Delta \epsilon \times \epsilon_d$
    \ELSE
        \item $\epsilon \gets \epsilon + \Delta \epsilon$
    \ENDIF

\ENDWHILE

\STATE Return traveled image $\mathrm{\mathbf{x}}^\prime$.

\vspace{1mm}
\end{algorithmic}
\label{alg:travel}
\end{algorithm}

% fig:calibration_2

% fig:fgsm-travel_2

\section{Representation space analysis}

\subsection{Shape of representation space}

\begin{figure}[!t]
    \centering
    \small

    \begin{tabular}{ccccc}
        &
        \hspace{3.5mm} ResNet50 &
        \hspace{3.5mm} MobileNetV2 &
        \hspace{3.5mm} ViT-B/16 &
        \hspace{3.5mm} Swin-T \\

        \rotatebox[origin=l]{90}{\hspace{13.0mm}$\mathrm{\mathbf{d}_{FGSM}}$} &
        \includegraphics[width=.20\linewidth]{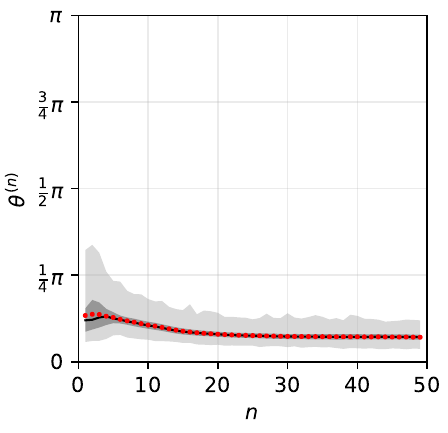} &
        \includegraphics[width=.20\linewidth]{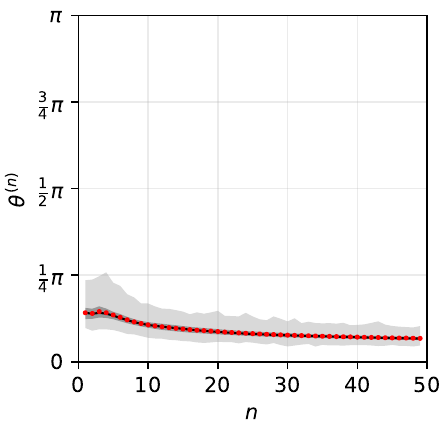} &
        \includegraphics[width=.20\linewidth]{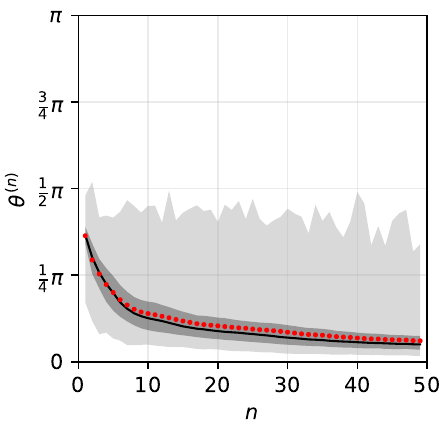} &
        \includegraphics[width=.20\linewidth]{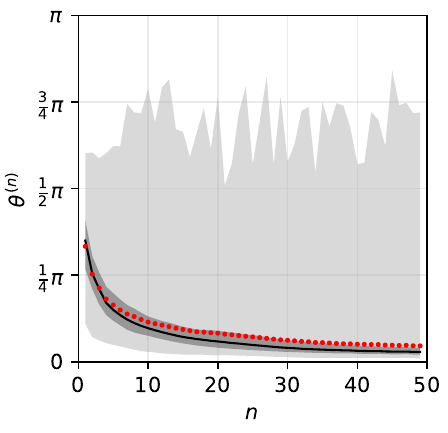} \\
        &
        \hspace{3.5mm} (a) &
        \hspace{3.5mm} (b) &
        \hspace{3.5mm} (c) &
        \hspace{3.5mm} (d)
    \end{tabular}

    \caption{
        Direction changes of output features with respect to steps of travel.
        \textbf{Light-gray regions}: Range between the minimum and maximum values.
        \textbf{Dark-gray regions}: Range between the first quartile (Q1) and the third quartile (Q3).
        \textbf{Black lines}: Medians (Q2).
        \textbf{Red dots}: Mean values.
    }\label{fig:footprint_around_csim}
\end{figure}

Fig.~\ref{fig:footprint_around_csim} shows the averaged direction changes in travel for 1000 different travel directions of 10 randomly chosen image data from the ImageNet validation set.
The directions are set using Eq.~\ref{eq:fgsm} with $y = 0, 1, \cdots, 999$ (1000 classes in ImageNet).
Large values of $\theta^{(1)}$ are observed for ViT-B/16 and Swin-T, indicating that the curvedness is not dependent on a particular choice of the travel direction.

\begin{figure}[!t]
    \centering
    \small

    \begin{tabular}{ccccc}
        &
        \hspace{4.0mm} ResNet50 &
        \hspace{4.0mm} MobileNetV2 &
        \hspace{4.0mm} ViT-B/16 &
        \hspace{4.0mm} Swin-T \\

        \rotatebox[origin=l]{90}{\hspace{13.0mm}$\mathrm{\mathbf{d}_{FGSM}}$} &
        \includegraphics[width=.20\linewidth]{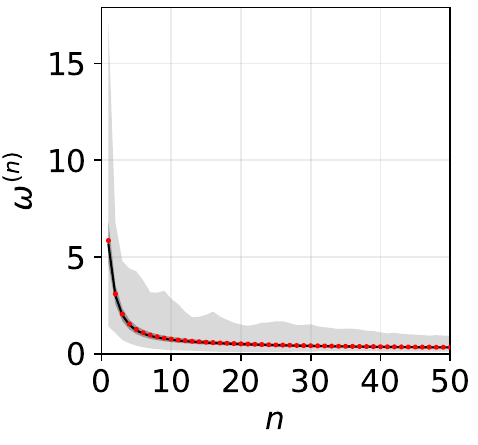} &
        \includegraphics[width=.20\linewidth]{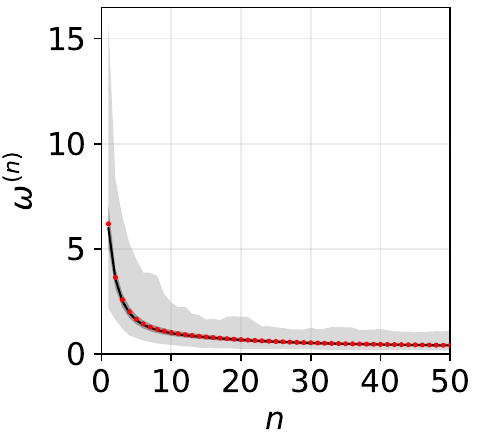} &
        \includegraphics[width=.209\linewidth]{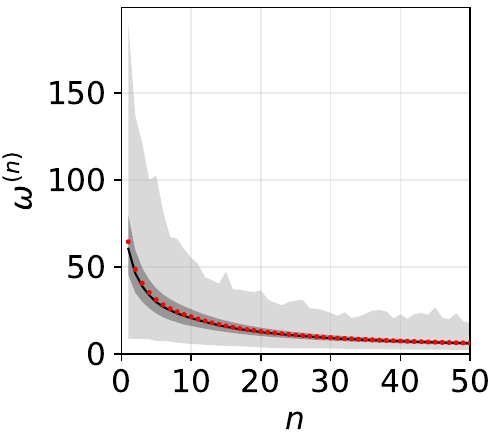} &
        \includegraphics[width=.20\linewidth]{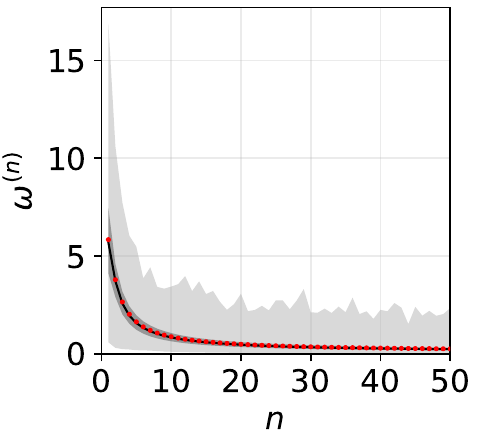} \\
        &
        \hspace{4.0mm} (a) &
        \hspace{4.0mm} (b) &
        \hspace{5.5mm} (c) &
        \hspace{4.0mm} (d) \\

        \rotatebox[origin=l]{90}{\hspace{11.0mm}$\mathrm{\mathbf{d}_{\mathbf{r}+FGSM}}$} &
        \includegraphics[width=.20\linewidth]{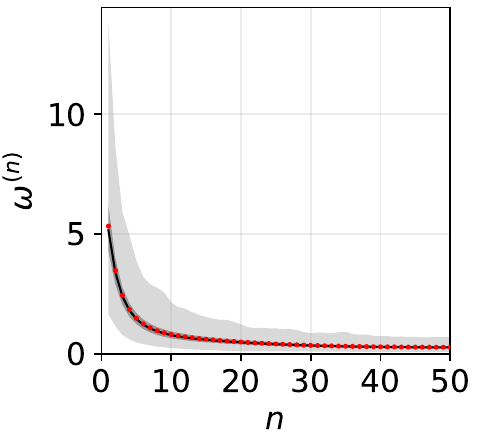} &
        \includegraphics[width=.20\linewidth]{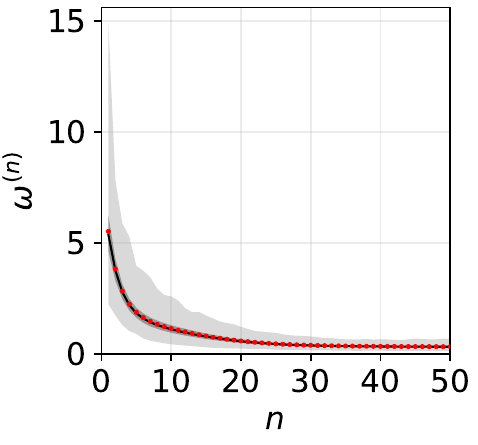} &
        \includegraphics[width=.209\linewidth]{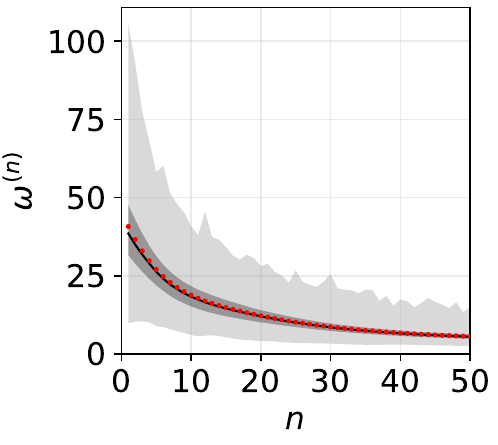} &
        \includegraphics[width=.20\linewidth]{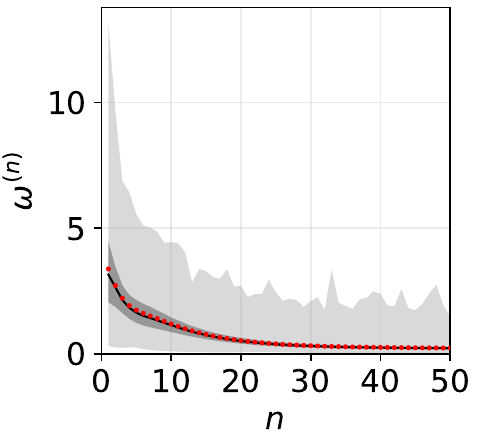} \\
        &
        \hspace{4.0mm} (e) &
        \hspace{4.0mm} (f) &
        \hspace{5.5mm} (g) &
        \hspace{4.0mm} (h)
    \end{tabular}

    \caption{
        Magnitude changes of output features with respect to steps of travel.
        \textbf{Light-gray regions}: Range between the minimum and maximum values.
        \textbf{Dark-gray regions}: Range between the first quartile (Q1) and the third quartile (Q3).
        \textbf{Black lines}: Medians (Q2).
        \textbf{Red dots}: Mean values.
    }\label{fig:footprint_cdist}
\end{figure}

Fig.~\ref{fig:footprint_cdist} shows the step-wise magnitudes in travel (i.e., $\omega^{(n)}$), which shows a similar trend to Fig.~\ref{fig:footprint_csim}.

% fig:footprint_around_csim

% cdist

\subsection{Validation of curves}

\begin{figure}[!t]
    \centering
    \small

    \begin{tabular}{ccccc}
        &
        \hspace{4.0mm} ResNet50 &
        \hspace{4.0mm} MobileNetV2 &
        \hspace{4.0mm} ViT-B/16 &
        \hspace{4.0mm} Swin-T \\

        \rotatebox[origin=l]{90}{\hspace{13.0mm}$\mathrm{\mathbf{d}_{FGSM}}$} &
        \includegraphics[width=.20\linewidth]{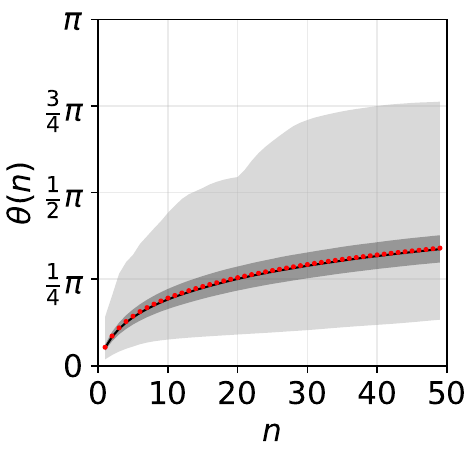} &
        \includegraphics[width=.20\linewidth]{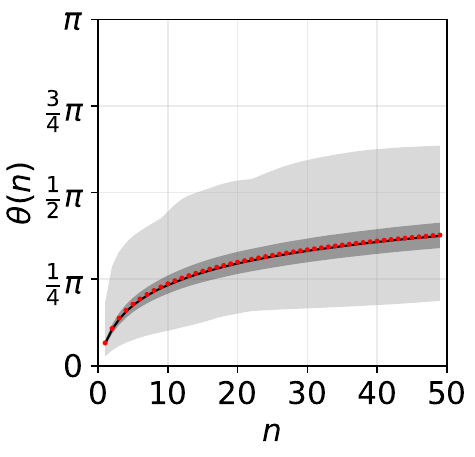} &
        \includegraphics[width=.20\linewidth]{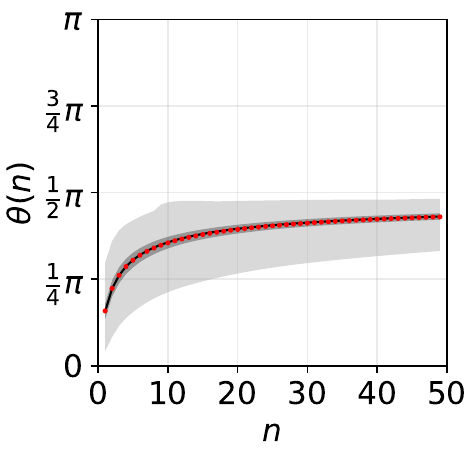} &
        \includegraphics[width=.20\linewidth]{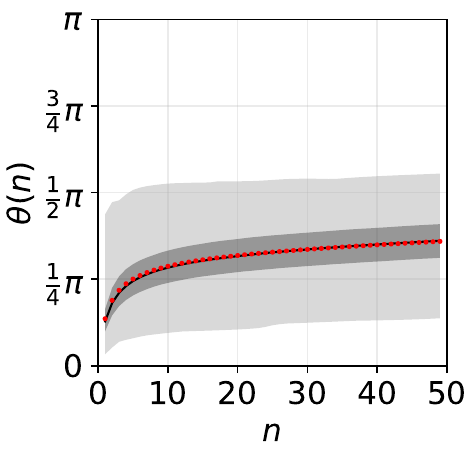} \\
        &
        \hspace{4.0mm} (a) &
        \hspace{4.0mm} (b) &
        \hspace{4.0mm} (c) &
        \hspace{4.0mm} (d)
    \end{tabular}

    \caption{
        Angle between the first movement and movement up to step $n$.
        \textbf{Light-gray regions}: Range between the minimum and maximum values.
        \textbf{Dark-gray regions}: Range between the first quartile (Q1) and the third quartile (Q3).
        \textbf{Black lines}: Medians (Q2).
        \textbf{Red dots}: Mean values.
    }\label{fig:footprint_sim}
\end{figure}

Fig.~\ref{fig:footprint_csim} shows that the input-output relationship of Transformers is locally nonlinear around the data.
Nonetheless, it could be contended that the travel may take a zigzag path, rather than a curved one.
To test such possibility, we define $\mathrm{\mathbf{z}}(n) \triangleq \mathrm{\mathbf{z}}^{(n)} - \mathrm{\mathbf{z}}^{(0)}$ ($n = 1, 2, \cdots, N$), which is the movement until step $n$ from the beginning.
Fig.~\ref{fig:footprint_sim} shows the angle between $\mathrm{\mathbf{z}}(1)$ and $\mathrm{\mathbf{z}}(n)$, which is denoted as $\theta(n)$.
If the travel follows a zigzag trajectory, $\theta(n)$ would oscillate.
However, we find that as $n$ increases, $\theta(n)$ tends to increase first and then becomes constant, which rejects the possibility of a zigzag path.

% sim

\subsection{Jump size}

We empirically observe that ViT-B/16 requires a bigger jump size ($\epsilon_{\mathrm{\mathbf{r}}}$) to escape from the curved region compared to Swin-T.
We hypothesize that this is due to the difference in the shape of the curved regions of ViTs and Swins.
Figs.~\ref{fig:footprint_csim}c and \ref{fig:footprint_csim}d show that $\theta^{(n)}$ decreases more slowly for ViT-B/16 than Swin-T, indicating wider curved regions in the former, i.e., ViT-B/16 requires a larger jump size.

\subsection{Measure of curvedness}

We observe that $\theta^{(1)}$ (the first direction change) and $\sum_{n=1}^{N-1}{\theta^{(n)}}$ (the total direction change) are highly correlated (correlation coefficients of 0.897, 0.904, 0.800, and 0.793 for ResNet50, MobileNetV2, ViT-B/16, and Swin-T, respectively).
Thus, we use $\theta^{(1)}$ for the measure of curvedness of the representation space.
In addition, it can provide the information about the local curvedness near $\mathrm{\mathbf{z}}_{0}$.

% visualize_jump
\begin{figure}[!t]
    \centering
    \small

    \includegraphics[width=0.30\linewidth]{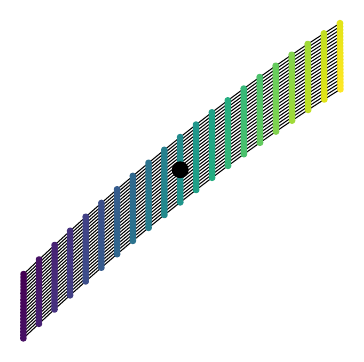}

    \caption{
        2D projected movements of the output in the representation space after random jump for Swin-T.
    }\label{fig:visualize_jump}
\end{figure}

\subsection{Output movement after jump}

Fig.~\ref{fig:visualize_jump} visualizes the 2D projected movement of the output of Swin-T in the representation space after random jump for the linear input movement in Fig.~\ref{fig:visualize}a.

\section{Nonlinearity of attention operation}

% fig:attn_data_dependent
\begin{figure}[!t]
    \centering
    \small

    \begin{tabular}{cc}
        \includegraphics[width=.35\linewidth]{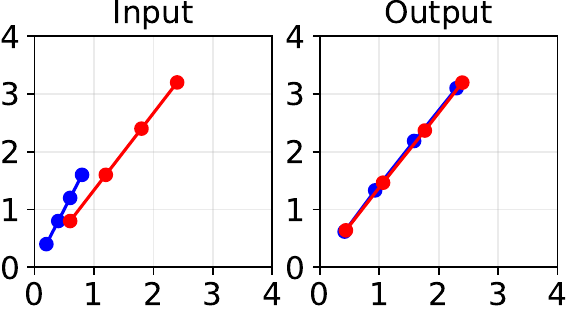} &
        \includegraphics[width=.35\linewidth]{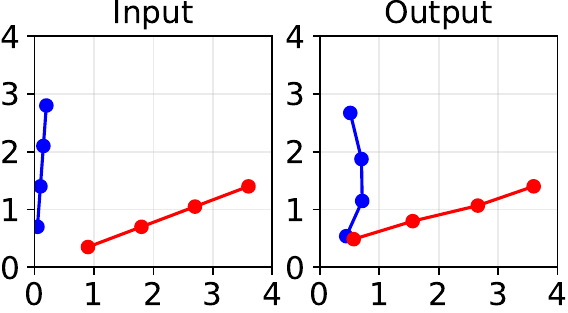} \\
        (a) Linear case &
        (b) Nonlinear case
    \end{tabular}

    \caption{
        An example of showing that the nonlinearity of the attention operation is data-dependent.
        Different colors indicate results of different tokens.
        The sequence of inter-connected points per token, progressing outward from the origin, indicates the gradual change of $\mathrm{\mathbf{P}}$ from $\mathrm{\mathbf{0}}$ to $\mathrm{\mathbf{X}}$, $2\mathrm{\mathbf{X}}$, and then $3\mathrm{\mathbf{X}}$ (on the left panels) or the corresponding change in the output of the attention operation (on the right panels).
    }\label{fig:attn_data_dependent}
\end{figure}

We present a simple example in low dimension to show that the nonlinearity of the attention operation depends on the input data, as shown in Eq.~\ref{eq:attn_moved_residual}.
Note that $\mathrm{\mathbf{X}}, \mathrm{\mathbf{P}} \in \mathbb{R}^{N_{t} \times D_{in}}$, $\mathrm{\mathbf{A}} \in \mathbb{R}^{N_{t} \times N_{t}}$, $\mathrm{\mathbf{W_{q}}}, \mathrm{\mathbf{W_{k}}} \in \mathbb{R}^{D_{in} \times D_{k}}$, and $\mathrm{\mathbf{W_{v}}} \in \mathbb{R}^{D_{k} \times D_{out}}$, where $N_{t}$ is the number of input tokens and $D_{in}$ and $D_{out}$ are the dimensions of input and output data, respectively.

Consider a simple case where $N_{t} = D_{in} = D_{k} = D_{out} = 2$, $\mathrm{\mathbf{W_{q}}}=\mathrm{\mathbf{W_{k}}}=\mathrm{\mathbf{W_{v}}}=\mathrm{\mathbf{I}}_{2}$, and $\mathrm{\mathbf{P}} \in \{\mathrm{\mathbf{0}}, \mathrm{\mathbf{X}}, 2\mathrm{\mathbf{X}}, 3\mathrm{\mathbf{X}}\}$.
For $\mathrm{\mathbf{X}} = \big[\begin{smallmatrix} 0.20 & 0.40 \\ 0.60 & 0.80 \end{smallmatrix}\big]$, $\mathrm{Attn}(\mathrm{\mathbf{X}} + \mathrm{\mathbf{P}})$ behaves as a linear operation (Fig.~\ref{fig:attn_data_dependent}a).
However, when $\mathrm{\mathbf{X}} = \big[\begin{smallmatrix} 0.05 & 0.70 \\ 0.90 & 0.35 \end{smallmatrix}\big]$, the operation becomes nonlinear (Fig.~\ref{fig:attn_data_dependent}b).

\section{Intriguing observations}

\subsection{Contribution of components to curvedness}

Table~\ref{tab:xray} shows the changes in $\theta^{(1)}$ when an input passes through each layer (component) in a model, denoted as $\Delta \theta^{(1)}$ (i.e., contribution of a component to curvedness).
The values (in radian) are first averaged for 50000 images of the ImageNet validation set, then averaged again within each kind of component.
The standard deviation values of the final averaged $\Delta \theta^{(1)}$ are also shown in the parentheses.
A value without standard deviation indicates that there is only a single component of the same kind in the model.
Hyperparameters for travel are $\epsilon/N$=.01 and $\mathrm{\mathbf{d}} = \mathrm{\mathbf{d}}_{\mathrm{FGSM}}$.

% tab:xray
\begin{table}[!t]
    \centering
    \small

    \caption{
        $\Delta \theta^{(1)}$ at each layer for major components in different models.
    }\label{tab:xray}

    \begin{tabular}{r|rrrrr}
        \multirow{2}{*}{} &
        \multicolumn{1}{c}{\multirow{2}{*}{ResNet50}} &
        \multicolumn{1}{c}{\multirow{2}{*}{Swin-T}} &
        \multicolumn{1}{c}{ConvNeXt} &
        \multicolumn{1}{c}{\multirow{2}{*}{DeiT-Ti}} &
        \multicolumn{1}{c}{DeiT-Ti} \\
        & & & \multicolumn{1}{c}{-Tiny} & & \multicolumn{1}{c}{-Distilled} \\
        \hline

        \multirow{2}{*}{Convolution}
        & $.000 \pi$ & $.000 \pi$ & $.000 \pi$ & $.000 \pi$ & $.000 \pi$ \\
        & ($\pm.000\pi$) & & ($\pm.000\pi$) & & \\

        \multirow{2}{*}{Self-Attention}
        & & $.224 \pi$ & & $.243 \pi$ & $.209 \pi$ \\
        & & ($\pm.116\pi$) & & ($\pm.118\pi$) & ($\pm.089\pi$) \\

        \multirow{2}{*}{BatchNorm}
        & $.000 \pi$ & & & & \\
        & ($\pm.000\pi$) & & & & \\

        \multirow{2}{*}{LayerNorm}
        & & $.125 \pi$ & $.111 \pi$  & $.124 \pi$  & $.105 \pi$  \\
        & & ($\pm.081\pi$) & ($\pm.029\pi$) & ($\pm.065\pi$) & ($\pm.051\pi$) \\

        \multirow{2}{*}{ReLU}
        & $.105 \pi$ & & & & \\
        & ($\pm.029\pi$) & & & & \\

        \multirow{2}{*}{GELU}
        & & $.241 \pi$ & $.236 \pi$ & $.254 \pi$ & $.227 \pi$ \\
        & & ($\pm.073\pi$) & ($\pm.058\pi$) & ($\pm.063\pi$) & ($\pm.054\pi$)
    \end{tabular}
\end{table}

\subsection{Different model behaviors}

\begin{figure}[!t]
    \centering
    \small

    \begin{tabular}{ccc}
        \includegraphics[width=.20\linewidth]{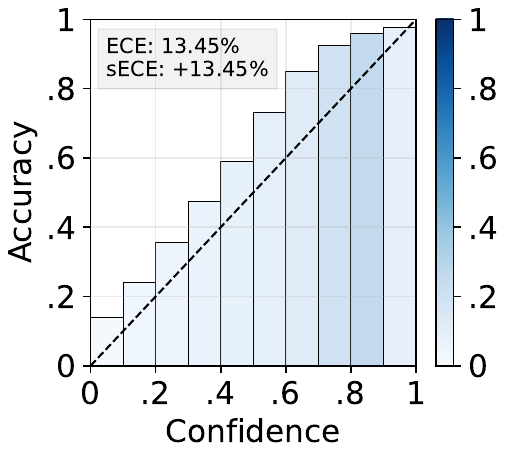} &
        \includegraphics[width=.20\linewidth]{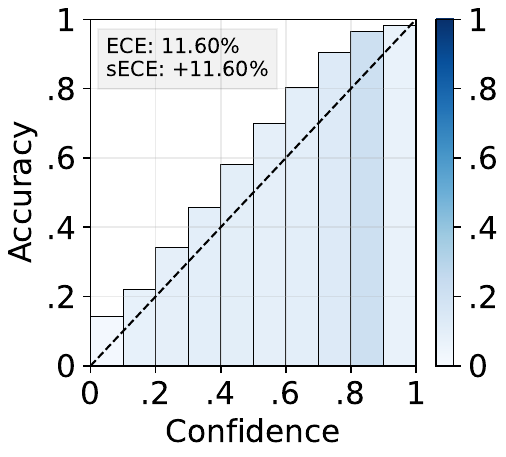} &
        \includegraphics[width=.20\linewidth]{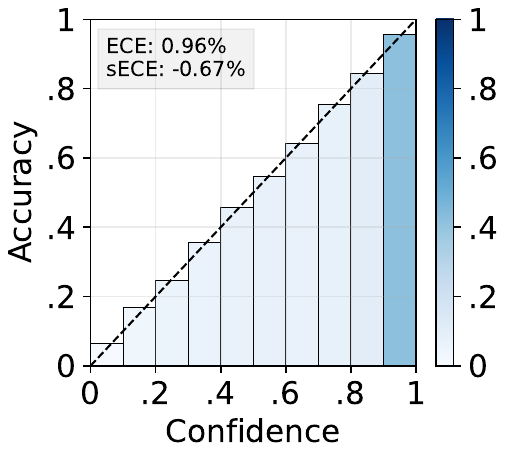} \\
        (a) ConvNeXt-Tiny &
        (b) DeiT-Ti &
        (c) DeiT-Ti-Distilled
    \end{tabular}

    \caption{
        Reliability diagrams of ConvNeXt and DeiT.
        Transparency of bars represent the ratio of the number of data in each confidence bin.
        ECE and sECE values are also shown in each case.
    }\label{fig:calibration_3}
\end{figure}

\begin{figure}[!t]
    \centering
    \small

    \begin{tabular}{ccc}
        \includegraphics[width=.20\linewidth]{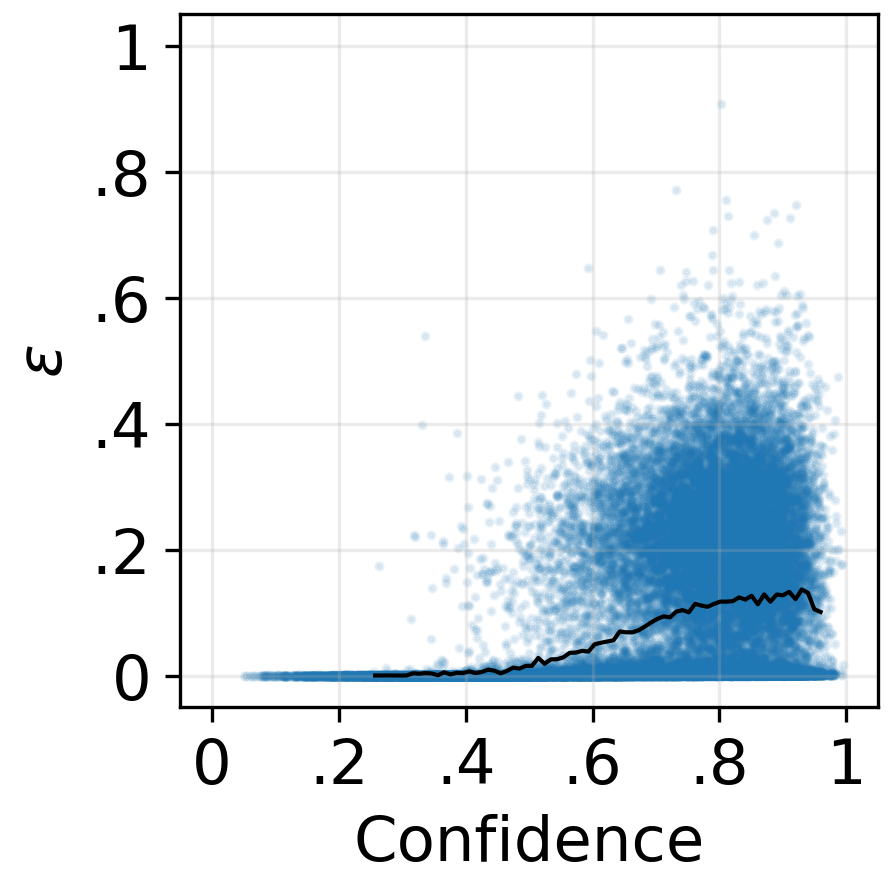} &
        \includegraphics[width=.20\linewidth]{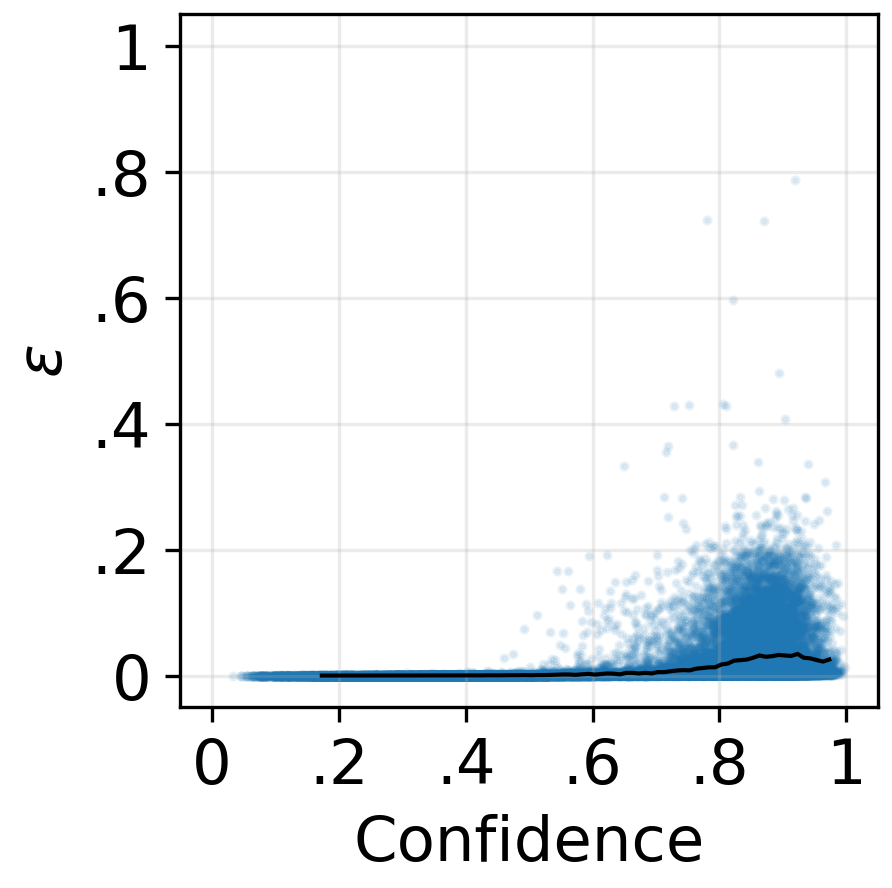} &
        \includegraphics[width=.20\linewidth]{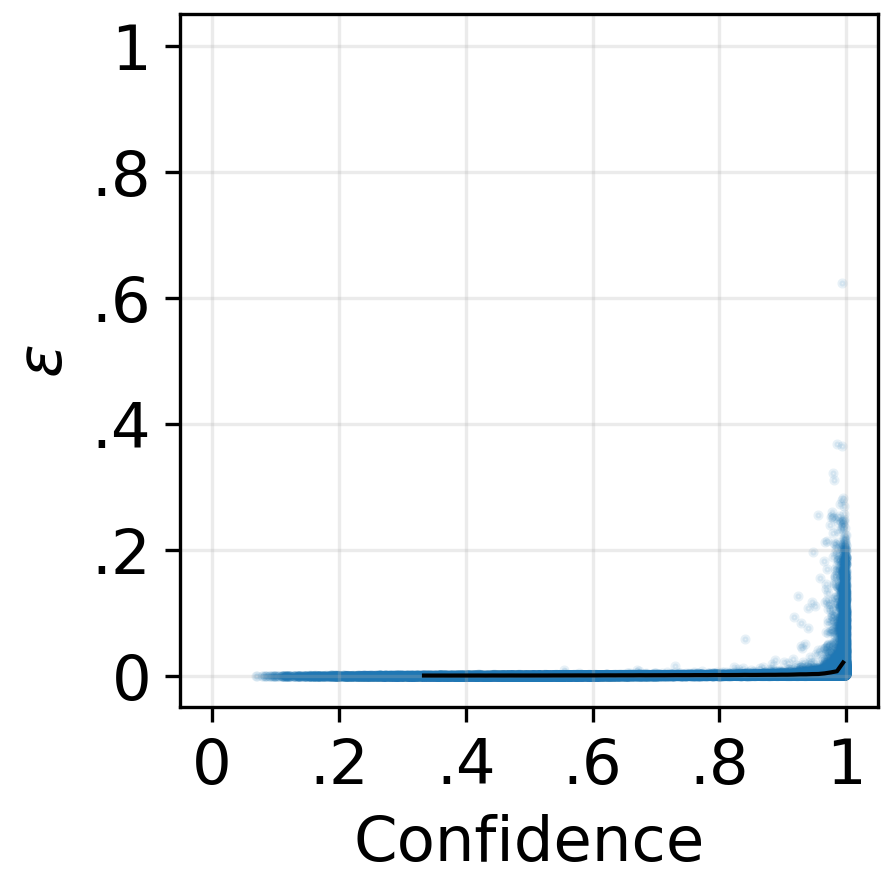} \\
        \hspace{5.0mm} (a) ConvNeXt-Tiny &
        \hspace{5.0mm} (b) DeiT-Ti &
        \hspace{5.0mm} (c) DeiT-Ti-Distilled
    \end{tabular}

    \caption{
        Lengths ($\epsilon$) of the travel to decision boundaries with respect to the confidence for ConvNeXt and DeiT.
        Black lines represent average values.
    }\label{fig:fgsm-travel_3}
\end{figure}

Figs.~\ref{fig:calibration_3} and \ref{fig:fgsm-travel_3} show the realibility diagrams and $\epsilon$ with respect to confidence, respectively, for ConvNeXt-Tiny, DeiT-Ti, and DeiT-Ti-Distilled.

% fig:calibration_3

% fig:fgsm-travel_3

\subsection{Curved space during training}

\begin{figure}[!t]
    \centering
    \small

    \includegraphics[width=.45\linewidth]{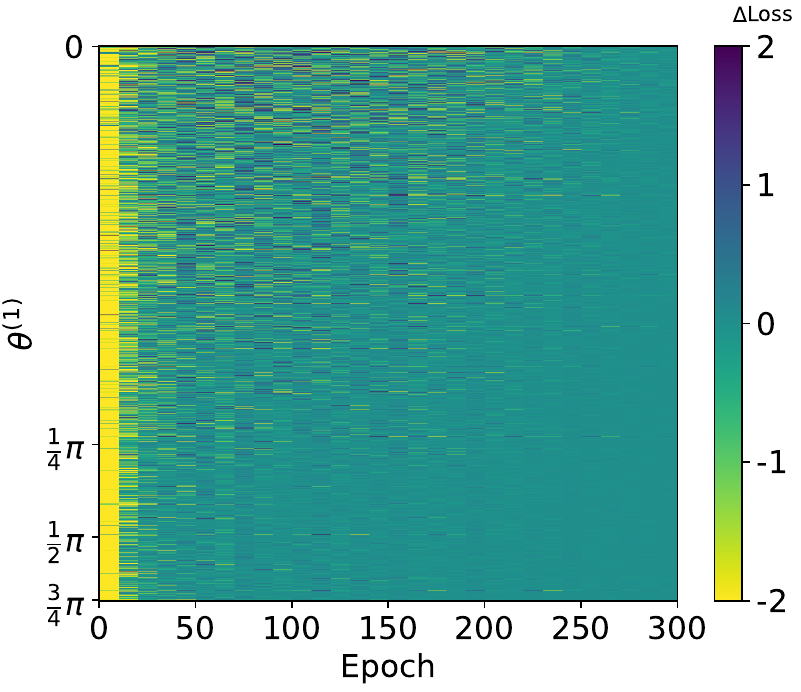}

    \caption{
        Loss changes of 10,000 training data from ImageNet during training of Swin-T.
        Each row represents the loss change at every 10 epochs for each data.
        The rows are sorted in an ascending order of $\theta^{(1)}$ after training.
        The loss change value is clipped within [-2,2] for visualization.
    }\label{fig:train_loss_change}
\end{figure}

Each row of Fig.~\ref{fig:train_loss_change} shows the loss change of each of the 10,000 ImageNet training data at every 10 epochs during training of Swin-T.
The rows are sorted in an ascending order of $\theta^{(1)}$ after training.

We follow the original training implementation code of the Swin-T for model training, e.g., 300 training epochs, batch size of 128, learning rate of $5\times 10^{-4}$, weight decay parameter of .05, and AdamW \cite{loshchilov2018decoupled} as the optimizer.

\begin{figure}[!t]
    \centering
    \small

    \begin{tabular}{ccc}
        \includegraphics[width=.20\linewidth]{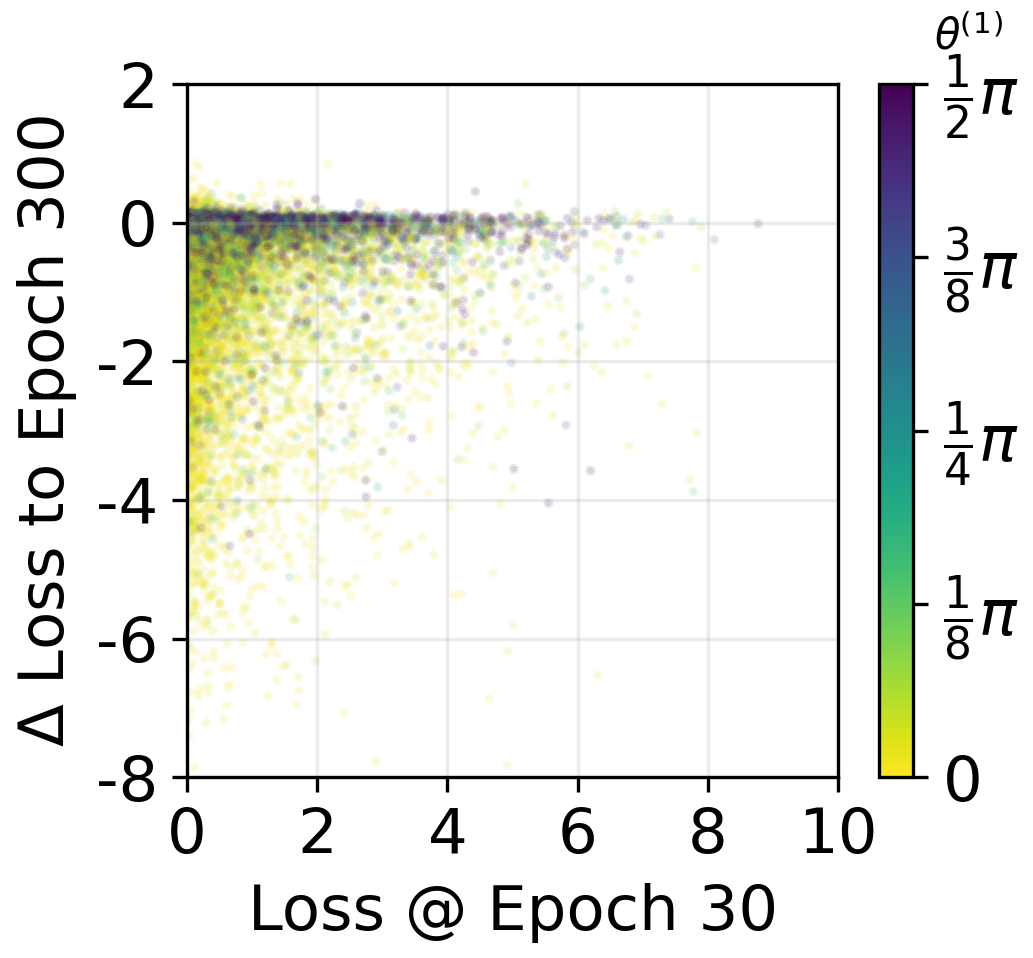} &
        \includegraphics[width=.20\linewidth]{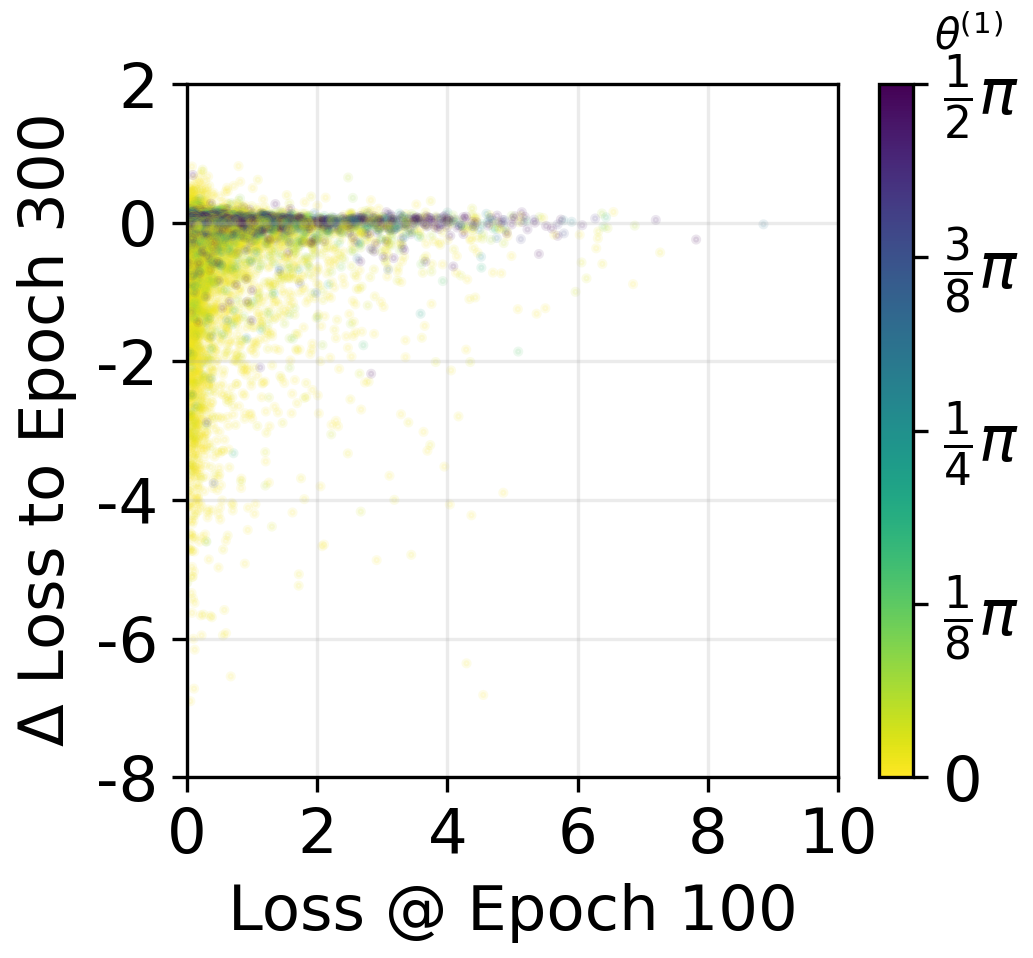} &
        \includegraphics[width=.20\linewidth]{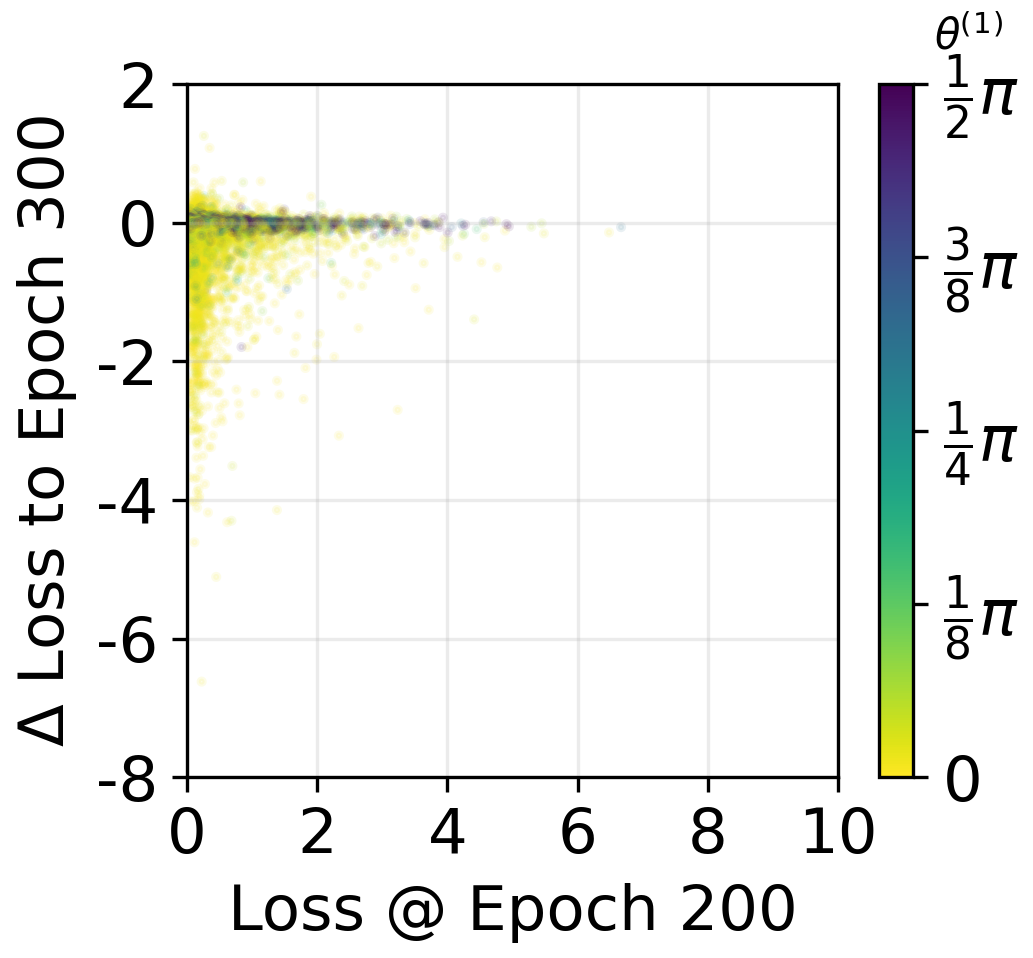}
    \end{tabular}

    \caption{
        Scatter plots showing the loss at a certain training epoch with respect to the loss change till the end of training for Swin-T.
        Colors indicate $\theta^{(1)}$ after training.
    }\label{fig:train_loss-loss}
\end{figure}

Fig.~\ref{fig:train_loss-loss} provides another view of Fig.~\ref{fig:train_loss_change} in terms of the relationship between the loss at a certain epoch and the loss change from the epoch until the end of training.

\begin{figure}[!t]
    \centering
    \small

    \begin{tabular}{ccc}
        \includegraphics[width=.20\linewidth]{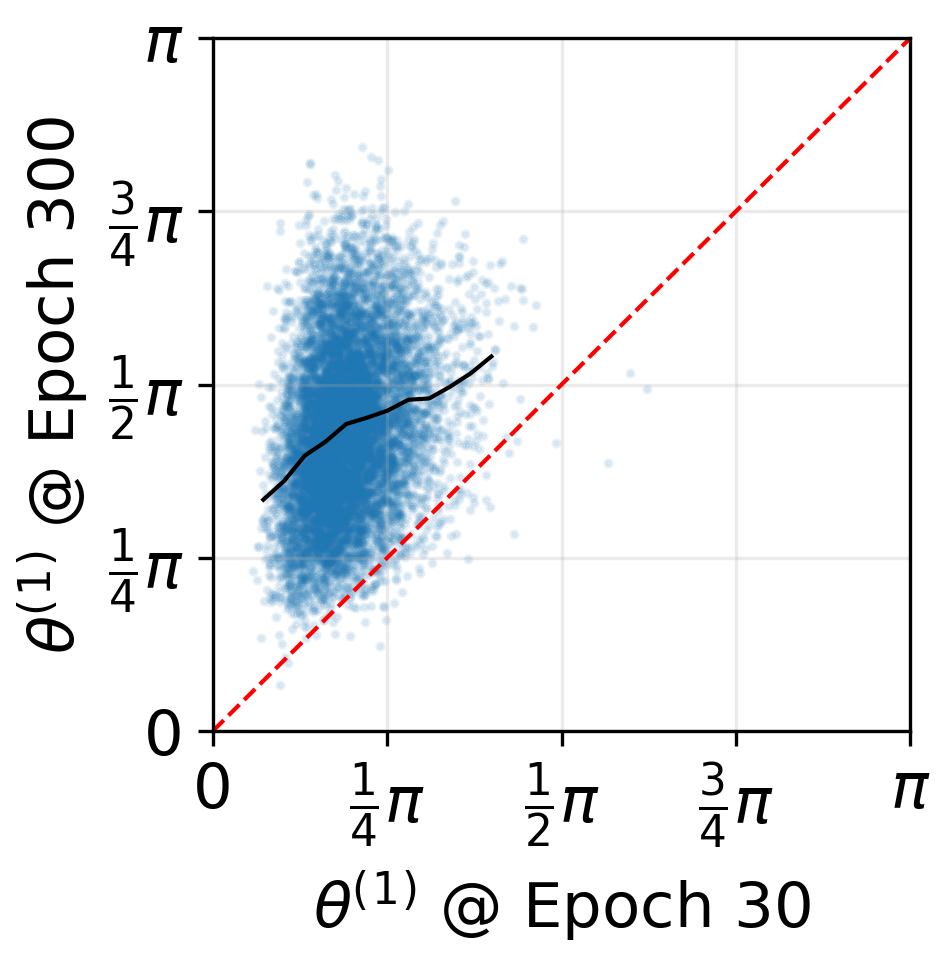} &
        \includegraphics[width=.20\linewidth]{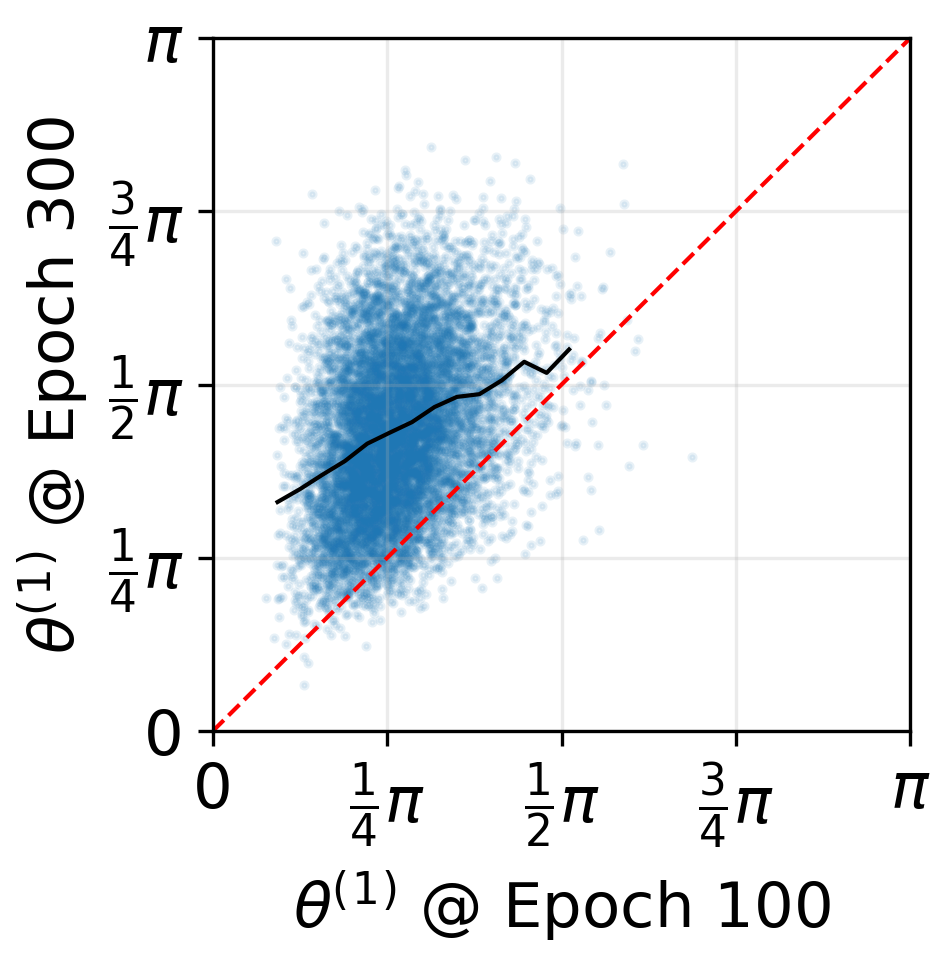} &
        \includegraphics[width=.20\linewidth]{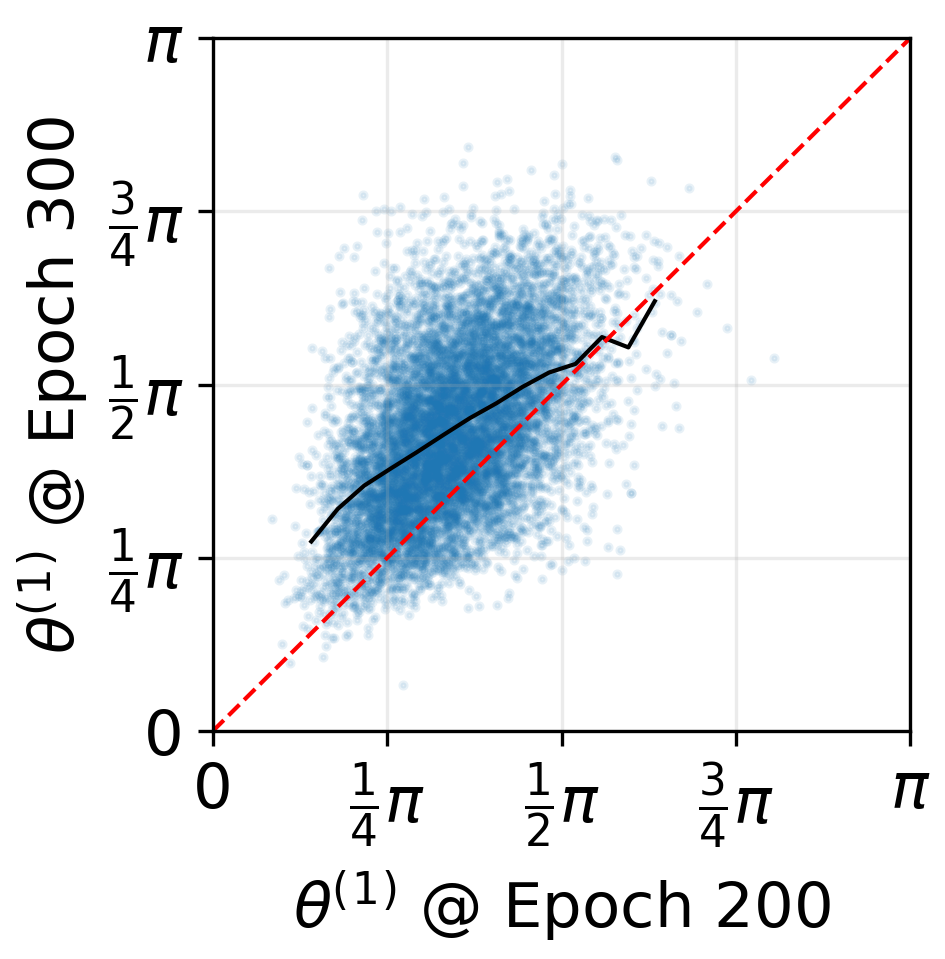}
    \end{tabular}

    \caption{
        Relationship of $\theta^{(1)}$ at a certain training stage and the final $\theta^{(1)}$ for Swin-T.
        Black lines indicate average values.
    }\label{fig:train_theta-theta}
\end{figure}

Fig.~\ref{fig:train_theta-theta} show the relationship of $\theta^{(1)}$ at a certain training stage and $\theta^{(1)}$ after training.

% fig:train_loss_change

% fig:train_loss-loss

% fig:train_theta-theta

\section{Pretrained models}

The models employed in this paper are pretrained models provided from PyTorch (\texttt{IMAGENET1K\_V1} as pretrained weights).
Specifically, we employ DeiT from its official GitHub repository, and use manually trained models in Section \textbf{Curved space during training}.

\end{document}